\documentclass[journal]{IEEEtran}

\usepackage{amsmath,amsfonts}
\usepackage{amsthm,amssymb}
\usepackage{mathrsfs}
\usepackage{algorithmic}
\usepackage{algorithm}
\usepackage{array}
\usepackage[caption=false,font=footnotesize]{subfig}
\usepackage{textcomp}
\IfFileExists{stfloats.sty}{\usepackage{stfloats}}{\usepackage{dblfloatfix}}
\usepackage{url}
\usepackage{graphicx}
\usepackage{cite}

\usepackage{bm}
\usepackage{booktabs}
\usepackage{multirow}
\usepackage{xcolor}

\usepackage{colortbl}
\newif\ifmarked
\ifdefined\MARKED \markedtrue \else \markedfalse \fi
\newif\ifprobe
\ifdefined\PROBE \probetrue \else \probefalse \fi

\newcommand{\rev}[1]{\ifprobe\textcolor[rgb]{0,0,1}{#1}\else#1\fi}
\newcommand{\revm}[1]{\ifprobe\begingroup\color[rgb]{0,0,1}#1\endgroup\else #1\fi}
\newcommand{\revrow}{\ifmarked\rowcolor{yellow}\fi}
\ifmarked
    \usepackage{eso-pic}
    
    \InputIfFileExists{main_marked_hl.tex}{}{\PackageWarningNoLine{revision}{main_marked_hl.tex not found; run make_marked.py}}
    \AddToShipoutPictureBG{\setlength{\unitlength}{1bp}\csname hlrects@\the\value{page}\endcsname}
\fi
\usepackage{eso-pic}
\usepackage[hidelinks]{hyperref}

\DeclareMathOperator*{\argmax}{arg\,max}
\DeclareMathOperator*{\argmin}{arg\,min}

\newcolumntype{C}[1]{>{\centering\arraybackslash}m{#1}}

\def\BibTeX{{\rm B\kern-.05em{\sc i\kern-.025em b}\kern-.08em
    T\kern-.1667em\lower.7ex\hbox{E}\kern-.125emX}}

\begin{document}

\title{LoRetta: A Foundation Model and Extensive Dataset for Global-Scale Remote Sensing Dense Image Matching}

\author{Siwei Yu, Han Guo, Zhenwei Shi,~\IEEEmembership{Senior Member,~IEEE},
    and Zhengxia Zou$^{*}$,~\IEEEmembership{Senior Member,~IEEE}%
    \thanks{This work was supported by the National Natural Science Foundation of China under Grant 62125102, Grant 62471014, Grant U24B20177, and Grant U25A20401, and in part by the Fundamental Research Funds for the Central Universities. (Corresponding author: Zhengxia Zou.)}%
    \thanks{Siwei Yu, Han Guo, Zhenwei Shi, and Zhengxia Zou are with the Department of Aerospace Intelligent Science and Technology, School of Astronautics, Beihang University, Beijing 100191, China, and also with the Key Laboratory of Spacecraft Design Optimization and Dynamic Simulation Technology (Beihang University), Ministry of Education (e-mail: \mbox{zhengxiazou@buaa.edu.cn}).}%
    \thanks{Project page: \url{https://siweiyu.com/work/loretta/}}}

\markboth{IEEE Transactions on Pattern Analysis and Machine Intelligence, 2026}%
{Yu \MakeLowercase{\textit{et al.}}: LoRetta: A Foundation Model and Extensive Dataset for Global-Scale Remote Sensing Dense Image Matching}

\IEEEaftertitletext{%
    \begin{center}
        \vspace{-0.5\baselineskip}
        \centering
        \includegraphics[width=\textwidth]{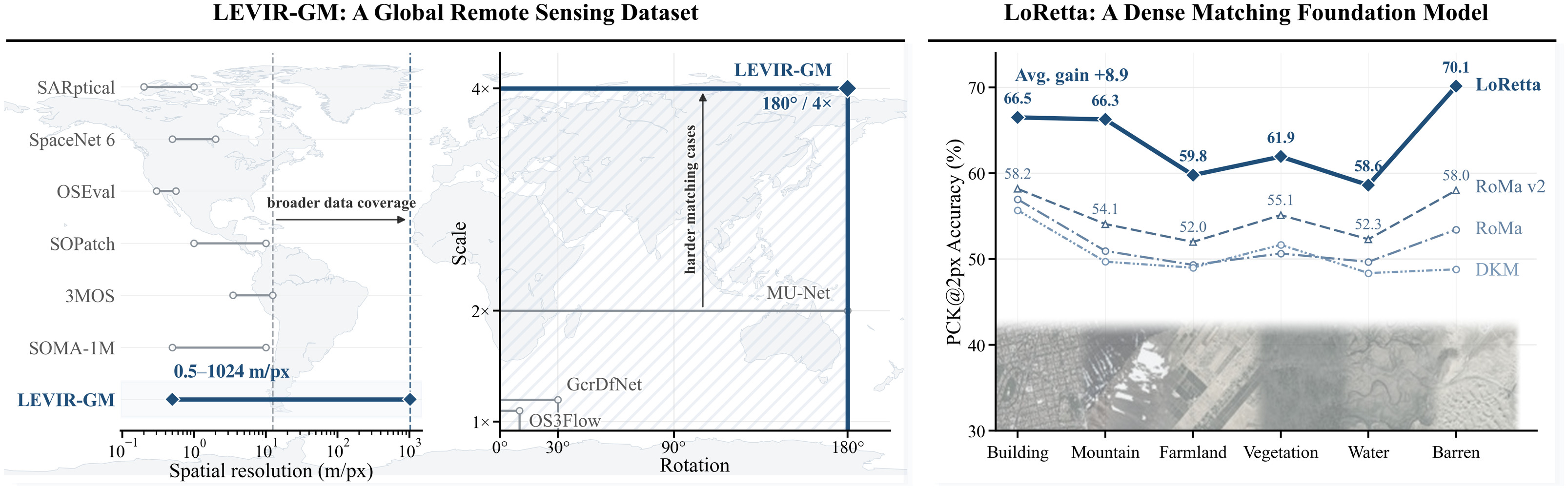}
        \vspace{-0.5\baselineskip}
        \refstepcounter{figure}
        \label{fig:intro_landscape}
        \parbox{\textwidth}{\footnotesize Fig.~\thefigure. \textbf{A global-scale dataset and foundation model for dense remote sensing image matching.} LEVIR-GM expands existing optical matching benchmarks in both spatial-resolution coverage and geometric variation, spanning sub-meter to kilometer-level imagery, global regions, full-angle rotations, and large scale differences. Built on this benchmark, LoRetta consistently improves the strict percentage of correct keypoints at 2 pixels (PCK@2px) over representative dense matchers across six land-cover groups, demonstrating stronger generalization to diverse global remote sensing scenes.}
        \vspace{0.5\baselineskip}
    \end{center}
}

\AddToShipoutPictureFG*{%
  \AtPageUpperLeft{\put(49,-171){\parbox[t]{514bp}{\centering\footnotesize Accepted manuscript of S. Yu, H. Guo, Z. Shi, and Z. Zou, ``LoRetta: A Foundation Model and Extensive Dataset for Global-Scale Remote Sensing Dense Image Matching,'' \textit{IEEE Transactions on Pattern Analysis and Machine Intelligence}, 2026. \url{https://doi.org/10.1109/TPAMI.2026.3759282}}}}%
  \AtPageLowerLeft{\put(49,27){\parbox[t]{514bp}{\raggedright\footnotesize \textcopyright\ 2026 IEEE. Personal use of this material is permitted. Permission from IEEE must be obtained for all other uses, in any current or future media, including reprinting/republishing this material for advertising or promotional purposes, creating new collective works, for resale or redistribution to servers or lists, or reuse of any copyrighted component of this work in other works.}}}%
}
\maketitle

\begin{abstract}
    Dense image matching establishes pixel-wise correspondences and underpins broad applications in computer vision and photogrammetry. However, extending dense matching to global-scale remote sensing remains challenging because image pairs may differ in acquisition time, season, viewpoint, spatial resolution, and land-cover state. The resulting large geometric offsets, partial overlap, and intrinsically unmatchable regions make direct dense correspondence prediction unreliable and inefficient.
    We thus reformulate dense matching as localization-and-registration: first localizing the matchable overlap and affine geometry, then refining dense residuals within the aligned frame. Based on this formulation, we propose LoRetta, a foundation model coupling matchability-aware affine localization with guided dense registration. We also introduce LEVIR-GM, a global-scale multi-temporal optical matching benchmark with dataset-native matchability labels (103K aligned, 827K augmented pairs, six continents, five years, 0.5--1024 m resolution). We further establish a unified evaluation protocol for sparse, semi-dense, and dense matchers.
    On LEVIR-GM, LoRetta achieves an area under the curve (AUC) of 83.3\%, outperforming the strongest baseline RoMa v2 by 1.6 points, with larger percentage of correct keypoints (PCK) gains of 6.5 and 8.2 points at 1 and 2 pixels, while reducing inference latency by 47.8\%. Astronaut-to-satellite and unmanned aerial vehicle (UAV)-to-satellite geolocalization \rev{and large-image registration} experiments further demonstrate its transferability as a reusable geometric aligner.
\end{abstract}


\section{Introduction}\label{sec:introduction}
\IEEEPARstart{D}{ense} image matching, which aims to establish pixel-wise correspondences between multiple views, is foundational in computer vision and photogrammetry. Unlike sparse matching that only associates discrete keypoints, dense matching provides a continuous and detailed representation of scene geometry. In remote sensing, it underpins a wide range of tasks, including image registration, change detection, disaster assessment, and global localization~\cite{xu2024localSurvey,li2025rsimSurvey}.

In recent years, deep learning has notably advanced dense image matching, especially in static natural scenes~\cite{truong2023pdcnetplus,edstedt2023dkm,edstedt2024roma,edstedt2025romav2}. However, global-scale remote sensing imagery is often acquired across years, seasons, viewpoints, and sensors, exhibiting extreme geometric deformations, radiometric differences, and complex land-cover changes. Consequently, existing dense matching methods struggle to generalize to remote sensing scenarios, where the assumptions of high overlap and consistent appearance break down.

For global-scale matching, we identify three challenges:

First, representative training datasets and benchmarks for global-scale scenarios are severely lacking. Most large-scale datasets focus on optical--synthetic aperture radar (SAR) alignment, with image pairs typically acquired within short temporal intervals~\cite{xu2023sopatch,schmitt2018sen12,ye20253mos,wu2026soma1m}. Consequently, multi-temporal optical-optical pairs, which exhibit far greater scene dynamics, remain rare. More importantly, such datasets rarely provide matchability labels, which are crucial for training robust matching models~\cite{lindenberger2023lightglue,sun2021loftr,edstedt2023dkm} but remain exceptionally scarce in remote sensing. Furthermore, existing benchmarks typically concentrate on high-resolution imagery finer than 10 m/pixel, leaving \rev{coarse-resolution} global observations underrepresented.

\begin{figure}[!t]
    \centering
    \includegraphics[width=\columnwidth]{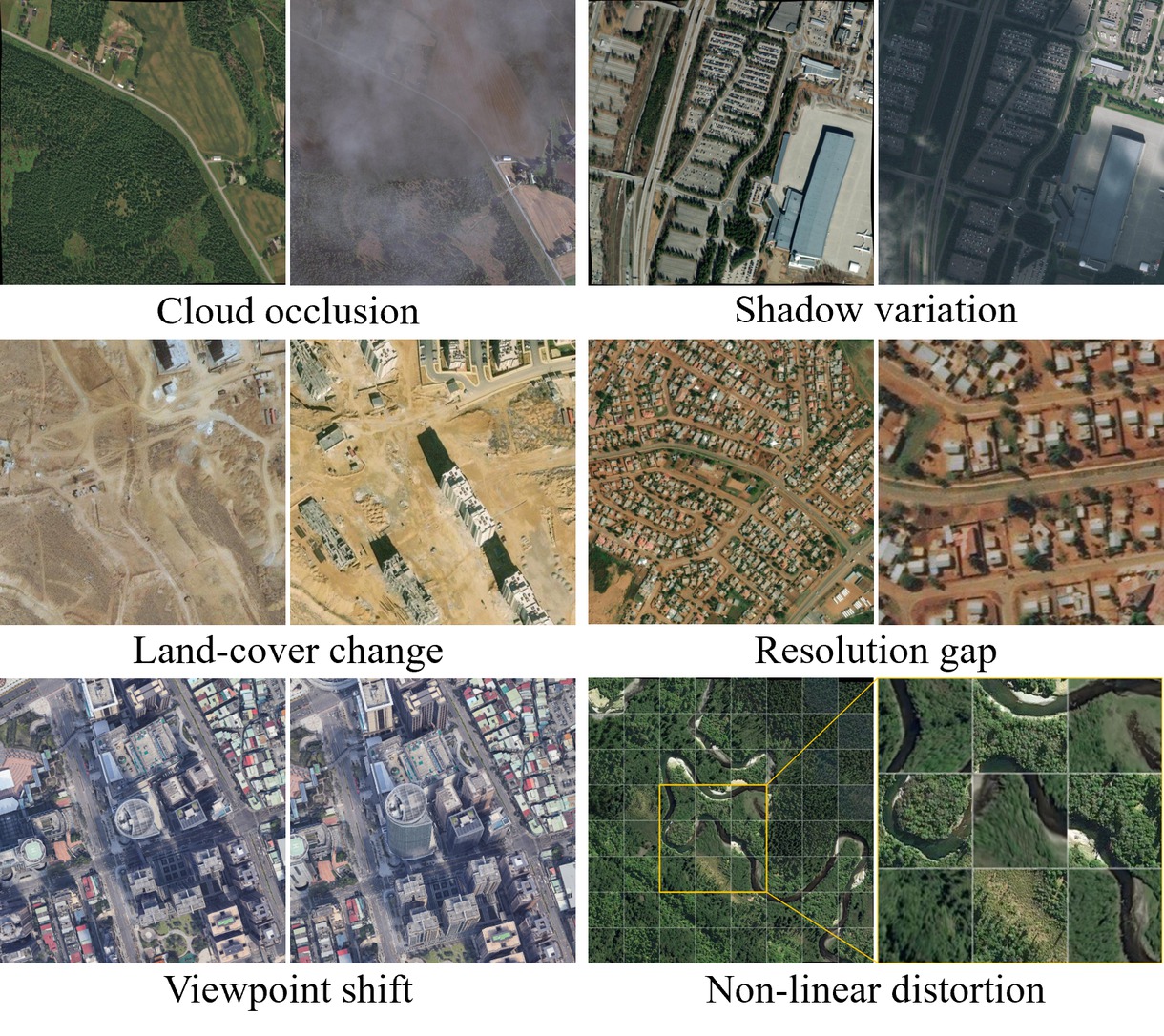}
    \caption{
        \textbf{Representative visual and geometric challenges in global-scale remote sensing matching.} Examples include cloud occlusion, shadow variation, land-cover change, resolution gap, viewpoint shift, and non-linear distortion.
    }
    \label{fig:global_matching_challenges}
\end{figure}

Second, global scenes are visually and geometrically complex, making reliable correspondence search difficult. As illustrated in Fig.~\ref{fig:global_matching_challenges}, multi-temporal imagery may exhibit cloud occlusion, shadow variation, and land-cover change, while differences in acquisition geometry and spatial resolution can introduce resolution gaps, viewpoint shifts, and local non-linear distortions. In particular, images with severe cloud occlusion are often excluded because reliable matched point pairs are difficult to obtain~\cite{schmitt2018sen12,ye20253mos}.

Third, dense matching is often too expensive for time-sensitive pipelines. On a single RTX 4090 GPU, sparse or semi-dense methods~\cite{lindenberger2023lightglue,sun2021loftr} often process a \(512\times512\) image pair in less than 35 ms, while recent state-of-the-art dense matchers~\cite{edstedt2024roma,edstedt2025romav2} require more than 100 ms per pair. This latency becomes a practical bottleneck when a system must verify many candidate images or repeatedly update its localization estimate.

These challenges motivate a different view of global-scale matching: rather than estimating dense correspondences directly between two globally unaligned images, the task should first localize the valid overlapping footprint and then register local residual geometry within the aligned frame. Built upon this view, we propose \textbf{LoRetta}, a dense matching foundation model with affine priors based on the ``\textbf{Lo}calization-and-\textbf{Re}gistration" paradigm, together with \textbf{LEVIR-GM}, an extensive dataset and benchmark designed for \textbf{G}lobal-scale remote sensing image \textbf{M}atching. To the best of our knowledge, LEVIR-GM is the first global-scale multi-temporal optical matching dataset of its kind. It comprises more than 800,000 \rev{image pairs}, spans six continents and five years (2018--2022), covers spatial resolutions from 0.5 m to 1024 m, and uses matchability labels to indicate where scene changes and matching difficulty make dense correspondences unreliable. Trained on this extensive data, LoRetta effectively handles complex earth observation scenes while ensuring high dense matching efficiency. A matchability-weighted affine locator first estimates a global affine prior from coarse correspondences, improving robustness to large rotations, scale changes, limited overlap, and unreliable regions. Conditioned on this affine-aligned frame, a localization-guided dense \rev{registration module} refines only local residual displacement, thereby reducing invalid global search and avoiding expensive dense computation over non-overlapping or transient areas. Training further uses pseudo-matchability maps, warp supervision on matchable regions, and affine supervision so that the model learns both where dense correspondences are reliable and how to stabilize global initialization. Fig.~\ref{fig:intro_landscape} summarizes the dataset coverage, geometric variation range, and land-cover-wise PCK@2px performance that motivate and validate the proposed LEVIR-GM and LoRetta.

On LEVIR-GM, we train LoRetta and fine-tune representative baselines~\cite{lowe2004sift,detone2018superpoint,lindenberger2023lightglue,sun2021loftr,edstedt2023dkm,edstedt2024roma,edstedt2025romav2}. LoRetta achieves 83.3\% AUC, outperforming the strongest baseline by 1.6 points, with notably larger gains at strict pixel-error thresholds, while reducing inference time by nearly half compared with the leading dense matcher. These gains are consistent across land-cover groups, geometric deformation ranges, and matchability conditions. In downstream astronaut-to-satellite and UAV-to-satellite localization \rev{and large-image registration}, LoRetta also transfers effectively to practical pipelines beyond the benchmark.

The contributions of this paper are summarized as follows:

\begin{itemize}
    \item
          We propose LoRetta, a foundation model for dense image matching in remote sensing based on a ``Localization-and-Registration'' paradigm. It first localizes the valid overlap with a matchability-weighted affine locator and then performs multi-scale dense residual registration, with geometric supervision on matchable regions, pseudo-matchability supervision, and affine supervision.

    \item
          We introduce LEVIR-GM, the first global-scale multi-temporal optical matching dataset with dataset-native matchability labels and presently the largest optical-optical remote sensing dense matching benchmark. It significantly expands the temporal span, spatial-resolution range, and geographic coverage required to train foundation models and evaluate image matching under severe land-cover and radiometric variations.

    \item
          We establish a unified dense-registration evaluation protocol for sparse, semi-dense, and dense matchers, and benchmark representative deep learning matching models in global remote sensing scenes. The results show that LoRetta raises the accuracy-efficiency upper bound on LEVIR-GM.
          We further evaluate LoRetta on \rev{various downstream applications, demonstrating its transferability beyond the benchmark}.

\end{itemize}

\section{Related work}

In this section, we briefly review the recent advancements in local feature matching, scalable training pipelines for image matching, and remote sensing image matching.

\subsection{Local Feature Matching}

Local feature matching is a fundamental cornerstone of computer vision. According to the density of established correspondences, existing deep learning-based methods can be broadly categorized into sparse, semi-sparse, semi-dense, and dense matching.

Sparse methods typically inherit the classical detect-describe-match pipeline~\cite{xu2024localSurvey}. While early handcrafted methods like SIFT~\cite{lowe2004sift} extract scale- and rotation-invariant keypoints, learning-based approaches such as SuperPoint~\cite{detone2018superpoint} and SuperGlue~\cite{sarlin2020superglue} advance this paradigm with self-supervised detection and graph-based attention matching, and LightGlue~\cite{lindenberger2023lightglue} further improves efficiency through adaptive depth and width. However, detector-based methods often fail in texture-less regions where reliable keypoints are scarce. To address this, detector-free semi-dense methods have emerged. LoFTR~\cite{sun2021loftr} pioneers Transformer-based coarse-to-fine matching, significantly improving performance in low-texture areas. Subsequent works like ASpanFormer~\cite{chen2022aspanformer} and Efficient LoFTR~\cite{wang2024efficientloftr} introduce flow-guided adaptive attention and aggregated attention mechanisms, respectively, to better capture local context and improve computational efficiency. More recently, to reduce the reliance on fixed spatial priors, RCM~\cite{lu2024rcm} and RCM+~\cite{lu2026freeform} introduce semi-sparse and free-form matching paradigms. By employing conflict-free matching strategies and position-agnostic encoding, they enable highly flexible zero-shot matching for arbitrary inputs. Despite their success, sparse methods remain dependent on keypoint repeatability, while semi-dense and semi-sparse variants are constrained by discretized correspondences, limiting their ability to capture pixel-wise details in complex scenes.

Dense matching aims to estimate correspondences for every pixel, often treating matching as a flow estimation or regression problem. PDC-Net~\cite{truong2021pdcnet} and its enhanced version PDC-Net+~\cite{truong2023pdcnetplus} introduce a probabilistic approach to jointly learn flow prediction and its uncertainty. DKM~\cite{edstedt2023dkm} decomposes dense matching into a kernel regression global matcher and a warp refinement process, significantly enhancing matching accuracy. Recently, RoMa~\cite{edstedt2024roma} and RoMa v2~\cite{edstedt2025romav2} incorporate visual foundation models~\cite{oquab2023dinov2,simeoni2025dinov3} for their robust feature representations, while UFM~\cite{zhang2025ufm} simplifies the architecture by directly regressing flow using a generic Transformer. These dense methods achieve high accuracy on standard benchmarks, but they typically rely on data augmentation to implicitly learn geometric invariance. In global remote sensing scenes, where images exhibit arbitrary rotations and large scale variations, the lack of explicit geometric modeling limits their performance.

Several works have integrated geometric priors into matching to address these limitations, but they typically apply priors locally~\cite{mishkin2018affnet} or rely on external geometric estimators~\cite{chang2023sem,zhang2026mesa}, rather than learning a global geometric initialization jointly with dense registration. In contrast, LoRetta integrates explicit affine priors into a hybrid dense matching architecture that couples Transformer-based coarse localization with convolutional multi-scale dense registration, effectively handling both the extreme geometric deformations and the computational bottlenecks.

\subsection{Scalable Training Pipelines for Image Matching}
\label{sec:training_pipelines}

Beyond architectural innovations, scaling up training data and supervision has become a key driver for robust image matching. GIM~\cite{shen2024gim} proposes a self-training framework that generates dense correspondence labels across video frames by combining complementary matchers and propagating filtered matches to distant frames, creating a self-improving loop for large-scale training without manual annotation. MINIMA~\cite{ren2025minima} addresses cross-modal generalization by using modality translation models to synthesize paired data (infrared, depth, event, etc.) from optical images, expanding the effective training distribution. MatchAnything~\cite{he2025matchanything} unifies these strategies, combining video propagation with generative augmentation to produce a training pipeline capable of handling diverse modalities and extreme viewpoint changes. While these training frameworks notably improve generalization, their data sources remain confined to internet videos and modality-translated imagery, which lack the global geographic diversity, multi-resolution characteristics, large-scale geometric deformations, and seasonal radiometric variations inherent to satellite remote sensing.

\subsection{Remote Sensing Image Matching}

Remote sensing image matching poses unique challenges compared to general computer vision due to large-scale geometric deformations, radiometric differences, and the complexity of earth's surface. Existing deep learning-based approaches can be broadly categorized into area-based, feature-based, and end-to-end regression methods~\cite{li2025rsimSurvey}.

Area-based and feature-based methods represent two classical paradigms. Area-based approaches locate correspondences by sliding a template window over the target image to maximize a predefined or learned similarity metric~\cite{wang2018deepregistration,zhang2022ddfn}. In contrast, feature-based methods sequentially detect keypoints, extract local descriptors, and establish matches through similarity matching and geometric verification~\cite{ye2018cnnfeatures,zhang2024micm}. Despite a certain tolerance to radiometric variations, both paradigms rely on isolated stages of feature extraction, matching, and outlier filtering, which may lead to cascading errors and often fail under the complex, repetitive patterns of global imagery.

Moving towards a more unified pipeline, end-to-end regression methods directly predict transformation parameters or dense flow fields. Several works regress global geometric transformations such as affine, homography, or projective parameters from image pairs~\cite{park2020twostream,li2023fusionregistration,chang2023gcrdfnet}. Others adopt iterative coarse-to-fine or cascaded strategies, progressively refining an initial global alignment through learned residual corrections~\cite{huang2021videosar,papadomanolaki2021multistep,ye2022munet,feng2022mid}. More recently, \rev{optical flow networks have been adapted to cross-modal registration, estimating pixel-wise flow for local relief distortions after coarse pre-registration or fitting a global affine transformation to reliable flow vectors under large geometric transformations}~\cite{zhang2023osflownet,sun2024os3flow,sun2025gdros}.

Despite these advancements, current remote sensing matching methods are mostly trained and evaluated on small-scale, region-specific datasets, lacking the capacity to generalize across diverse global scenes and seasons. Moreover, their matchability estimation often relies on hand-crafted constraints or simple symmetry checks, which struggle to adapt to unpredictable complex changes such as cloud cover and land use change.

\section{METHODOLOGY}

This section presents LoRetta in detail. We first motivate why an affine prior is a suitable macroscopic geometric model for multi-temporal earth observation. We then introduce the overall architecture, followed by the Matchability-Weighted Affine Localization and the Localization-Guided Dense Registration. Finally, we summarize the training objectives that connect dense warp-field estimation, matchability prediction, and affine supervision.

\subsection{Motivation: The Geometric Essence of Earth Observation and Affine Priors}

\begin{figure}[t]
    \centering
    \includegraphics[width=\linewidth]{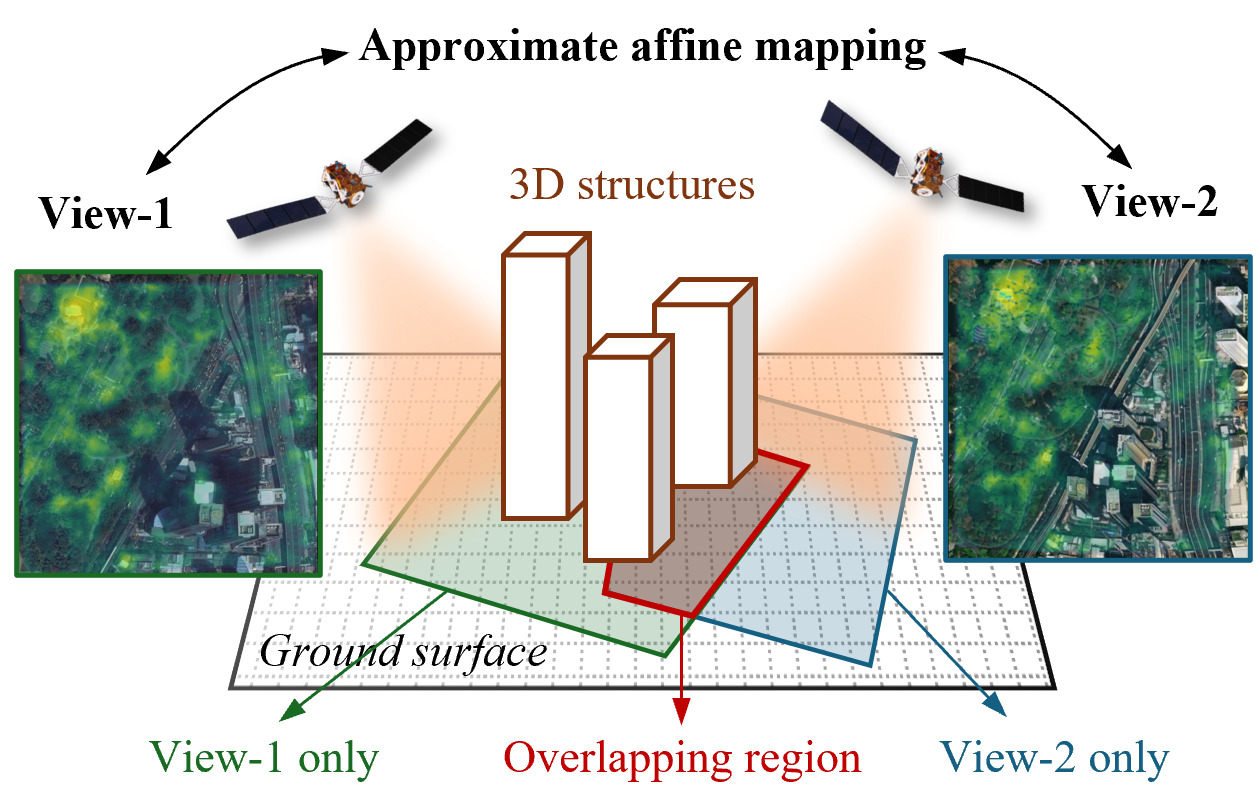}
    \caption{\textbf{Geometric motivation for the affine prior.} Two satellite observations may contain large non-overlapping footprints, shown by the green and blue regions, and a valid overlapping region that requires registration, shown in red. Within the overlap, the dominant ground surface can be approximated by an affine mapping under weak-perspective imaging, whereas above-ground 3D structures may introduce parallax, occlusion, or temporal changes.}
    \label{fig:motivation}
\end{figure}

Unlike object-centric natural scenes where dense matchers assume substantial overlap and stable appearance, earth observation imagery introduces a fundamentally different geometric structure. As illustrated in Fig.~\ref{fig:motivation}, two satellite observations of a common geographic region may present two geometric challenges. First, large footprint offsets leave substantial non-overlapping areas, shown by the green and blue regions. A global dense correspondence search over these independent regions can introduce irrelevant correlations and wrong matches. Second, even within the valid overlapping footprint, shown in red, the scene contains both a dominant ground-surface mapping and view-dependent 3D structures. The continuous ground surface is the primary target of remote sensing registration, whereas buildings and other above-ground objects can exhibit parallax, occlusion, or temporal changes. Forcing pixel-wise correspondences on such unreliable regions is therefore geometrically ill-posed.

For global-scale multi-temporal dense matching, the registration problem can therefore be framed as first localizing the valid overlapping footprint and then aligning the dominant macro-geographic ground surface. Because satellite imaging is acquired from a much greater distance than the observed terrain relief, the mapping of a locally planar ground surface between two observations can be modeled under a weak-perspective approximation. This yields an affine transformation that captures the dominant translation, rotation, scale, and shear before local residual distortions are refined.

This geometric insight motivates the proposed ``Localization-and-Registration'' paradigm. Rather than directly predicting an unconstrained global dense warp, LoRetta first estimates an explicit affine prior with a matchability-weighted affine localization module. This prior aligns the reference image to the sensed-image frame and localizes the dominant overlapping region. Conditioned on this affine initialization, a localization-guided dense registration module estimates local residual deformations caused by terrain relief, viewpoint variation, and imaging perturbations, while reducing the influence of unreliable parallax-affected or non-overlapping regions.

\subsection{Overall Architecture of LoRetta}

\begin{figure*}[t]
    \centering
    \includegraphics[width=1.0\linewidth]{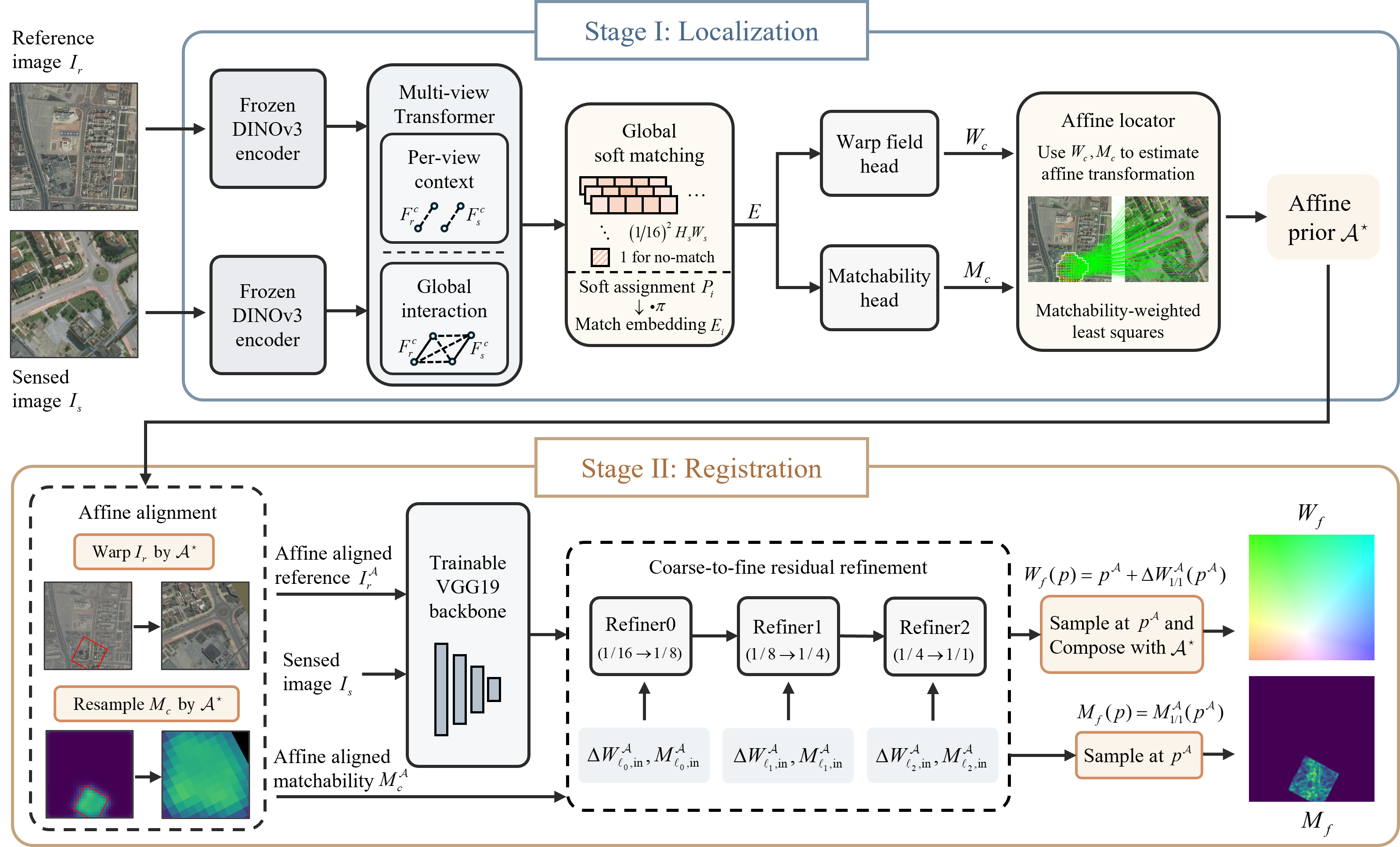}
    \caption{
        \textbf{Architecture of LoRetta.}
        \protect\rev{Stage~I contextualizes frozen DINOv3 descriptors at $1/16$ resolution with a multi-view Transformer, soft-matches them globally with a no-match candidate to predict the coarse warp $W_c$ and matchability $M_c$, and fits $\mathcal{A}^\star$ by matchability-weighted least squares.
        Stage~II warps $I_r$ and $M_c$ by $\mathcal{A}^\star$, and Refiner0--2 refine the residual $\Delta W^{\mathcal{A}}$ and matchability $M^{\mathcal{A}}$ from VGG19 features at $1/16$, $1/8$, and $1/4$ resolution, where $(a\to b)$ means refined at $a$ and interpolated to $b$.
        The fine outputs $W_f$ and $M_f$ sample the full-resolution maps at the affine-mapped coordinate $p^{\mathcal{A}}$.}
    }
    \label{fig:architecture}
\end{figure*}

As illustrated in Fig.~\ref{fig:architecture}, LoRetta takes a reference image and a sensed image as input and decomposes dense matching into two coupled stages. The localization stage predicts a coarse warp field and a matchability map, from which an affine prior is fitted using reliable coarse correspondences. This prior warps the reference image into the sensed-image frame and provides an aligned initialization for the registration stage, which then estimates residual dense displacements and the final matchability map.

The localization branch uses a frozen DINOv3~\cite{simeoni2025dinov3} feature extractor followed by a multi-view Transformer. For each input image, we extract intermediate DINOv3 features and concatenate multi-layer features to form coarse descriptors. The multi-view Transformer models within-image context and cross-image interactions to produce contextualized coarse descriptors. Global correlation and soft assignment then produce match embeddings, which are decoded into the coarse warp field and matchability map for affine fitting.

The registration branch performs trainable dense refinement on the affine-aligned image pair. The estimated affine transformation warps the reference image and resamples the coarse predictions into the sensed-image frame, yielding an affine-initialized registration problem. A VGG19-based~\cite{simonyan2015vgg} feature pyramid is constructed from the affine-warped reference image and the sensed image. Three learnable refinement modules, denoted as Refiner0, Refiner1, and Refiner2 in Fig.~\ref{fig:architecture}, then progressively estimate residual displacement fields from coarse to fine. The branch outputs refined dense warps and matchability maps, so that unreliable non-overlapping or transient regions can be identified rather than forced into dense correspondence.

\subsection{Matchability-Weighted Affine Localization}

Let \(I_r\) and \(I_s\) denote the reference and sensed images, respectively. We extract frozen DINOv3 coarse descriptors \(F_r^c\) and \(F_s^c\) at \(1/16\) resolution. The descriptors are processed by the multi-view Transformer in Fig.~\ref{fig:architecture}, which first performs per-view contextual encoding and then global cross-view interaction. The resulting contextualized descriptors encode both within-image structural context and cross-image geometric consistency before explicit affine fitting.

Given the contextualized coarse descriptors, we compute global cross-view correlation logits between every reference token and all sensed-image tokens. At \(1/16\) resolution, the sensed image provides \(N_s=(1/16)^2H_sW_s\) spatial candidates. We additionally append one learnable no-match candidate. A row-wise softmax over these \(N_s+1\) candidates yields \(P\in[0,1]^{N_r\times(N_s+1)}\), where \(P_{ij}\) denotes the soft assignment from reference token \(i\) to sensed token \(j\), and \(P_{i,\varnothing}\) denotes its no-match probability.

The soft-assignment distribution is further converted into a compact match embedding. Let \(\pi_j\) denote the positional embedding of sensed token \(j\), \(\pi_{\varnothing}\) the learnable no-match embedding, and \(E_i\) the resulting embedding for reference token \(i\), all in the same embedding space:
\begin{equation}
    E_i
    =
    \sum_{j=1}^{N_s}P_{ij}\pi_j
    +P_{i,\varnothing}\pi_{\varnothing}.
\end{equation}

Therefore, \(E_i\) encodes the softly assigned target position of reference token \(i\), while explicitly retaining the possibility that the token is unmatchable. Let \(E=\{E_i\}_{i=1}^{N_r}\). Separate warp-field and
matchability heads, jointly denoted by \(D(\cdot)\), decode the match embeddings into the coarse predictions:
\begin{equation}
    (W_c, M_c) = D(E).
\end{equation}
Here and throughout, \(W\) denotes a warp field storing normalized reference-to-sensed target coordinates, \(\Delta W\) denotes a residual displacement field in the same normalized coordinate system, and \(M\) denotes a matchability map.

The affine locator estimates a global affine prior from the normalized coordinates stored in \(W_c\) by matchability-weighted least squares. Let \(p_i\in[-1,1]^2\) denote the normalized center coordinate of the \(i\)-th reference patch, \(\tilde p_i=[p_i^\top,1]^\top\) its homogeneous coordinate, \(q_i=W_c(p_i)\in[-1,1]^2\) the corresponding target coordinate, and \(m_i=M_c(p_i)\in[0,1]\) its coarse matchability score. Given a matchability threshold \(\tau_A\), we fit the affine transform \(\mathcal A^\star\) from the selected coarse correspondences indexed by \(\Omega_A=\{i\mid m_i>\tau_A\}\):
\begin{equation}
    \mathcal A^\star
    =
    \argmin_{\mathcal A\in\mathbb R^{2\times3}}
    \sum_{i\in\Omega_A}
    m_i\|q_i-\mathcal A\tilde p_i\|_2^2.
\end{equation}
This formulation derives the affine prior directly from coarse correspondences weighted by predicted matchability, rather than regressing the six affine parameters.

\rev{After estimating the affine transform, we place the reference image and coarse matchability map into the sensed-image coordinate frame using $\operatorname{warp}(X;\mathcal{A}^\star)$, which samples $X$ backward through the inverse of $\mathcal{A}^\star$, bilinearly for the image and by nearest neighbor for matchability:}
\begin{equation}
    I_r^\mathcal{A} = \operatorname{warp}(I_r;\mathcal{A}^{\star}),
    \qquad
    M_c^\mathcal{A} = \operatorname{warp}(M_c;\mathcal{A}^{\star}).
\end{equation}
Here, \(I_r^{\mathcal{A}}\) is the affine-warped reference image, and \(M_c^{\mathcal{A}}\) is the coarse matchability map resampled in the affine-aligned frame. They provide the aligned image and reliability prior for subsequent residual registration.

\subsection{Localization-Guided Dense Registration}

After affine localization, LoRetta refines dense correspondences in the affine-aligned frame. The affine-warped reference image \(I_r^{\mathcal{A}}\) and the sensed image \(I_s\) are fed into a trainable VGG19 backbone~\cite{simonyan2015vgg}, denoted by \(\Phi_{\mathrm{VGG}}\), to extract a compact feature pyramid. We progressively refine the residual field at \(\ell_k\in\{1/16,1/8,1/4\}\), with \(k=0,1,2\), corresponding to Refiner0, Refiner1, and Refiner2:
\begin{equation}
    \{F_{r,\ell_k}^\mathcal{A},F_{s,\ell_k}\}_{k=0}^{2}
    = \Phi_{\mathrm{VGG}}(I_r^\mathcal{A},I_s).
\end{equation}
In this pyramid, \(F_{r,\ell_k}^\mathcal{A}\) and \(F_{s,\ell_k}\) denote the reference and sensed features extracted from \(I_r^{\mathcal{A}}\) and \(I_s\) at resolution \(\ell_k\), respectively. Each refiner updates the residual displacement field and matchability map at its own resolution.

At the coarsest aligned level \(\ell_0=1/16\), the affine transform provides the global coordinate initialization. We therefore initialize the inputs to the first refiner with zero residual displacement and the affine-warped coarse matchability map:
\begin{equation}
    \Delta W_{\ell_0,\mathrm{in}}^{\mathcal{A}}=0,
    \qquad
    M_{\ell_0,\mathrm{in}}^{\mathcal{A}}=M_c^{\mathcal{A}}.
\end{equation}
Starting from this affine-aligned initialization, the registration branch predicts only the residual displacement field that remains after macro-alignment, instead of re-estimating the full reference-to-sensed warp from scratch. This residual formulation reduces the search range for local refinement and makes the prediction target better conditioned under limited overlap or strong appearance variation.

At each level \(\ell_k\), \rev{the $k$-th refiner $R_k$, inspired by DKM~\cite{edstedt2023dkm}, refines the residual and matchability as}
\begin{subequations}
    \begin{align}
        F_{s,\ell_k}'(x)
         & =
        \revm{F_{s,\ell_k}(x'),
        \quad
        x'=x+\Delta W_{\ell_k,\mathrm{in}}^{\mathcal{A}}(x)}, \\
        \revm{C_{\ell_k}(x,\delta)}
         & =
        \revm{\bigl\langle
        F_{r,\ell_k}^{\mathcal{A}}(x),
        F_{s,\ell_k}(x'+\delta)
        \bigr\rangle/\sqrt{d_k}}, \\
        \Delta W_{\ell_k}^{\mathcal{A}},
        M_{\ell_k}^{\mathcal{A}}
         & =
        R_k\bigl(
        \revm{F_{r,\ell_k}^{\mathcal{A}},
        F_{s,\ell_k}',}
        C_{\ell_k},
        \Delta W_{\ell_k,\mathrm{in}}^{\mathcal{A}},
        M_{\ell_k,\mathrm{in}}^{\mathcal{A}}
        \bigr),
    \end{align}
\end{subequations}
\rev{where $x$ is a normalized grid coordinate at level $\ell_k$, $\delta$ ranges over the $(2r_k+1)^2$ grid-cell offsets within radius $r_k$, and $d_k$ is the feature dimension. $R_k$ is a convolutional residual network on the concatenated inputs that outputs the refined residual and, via a sigmoid, the matchability.}

For \(k<2\), the predictions are propagated to the next level using bilinear interpolation \(\mathcal U_k^{\mathrm{bil}}\) for the residual field, without rescaling its normalized displacement values, and nearest-neighbor interpolation \(\mathcal U_k^{\mathrm{nn}}\) for the matchability map:
\begin{equation}
    \begin{aligned}
        \Delta W_{\ell_{k+1},\mathrm{in}}^{\mathcal{A}}
         & =
        \mathcal U_k^{\mathrm{bil}}
        \!\left(\Delta W_{\ell_k}^{\mathcal{A}}\right), \\
        M_{\ell_{k+1},\mathrm{in}}^{\mathcal{A}}
         & =
        \mathcal U_k^{\mathrm{nn}}
        \!\left(M_{\ell_k}^{\mathcal{A}}\right).
    \end{aligned}
\end{equation}
\rev{Bilinear interpolation keeps the residual field continuous and sub-pixel, without staircase jumps, whereas nearest-neighbor interpolation copies each predicted matchability value rather than averaging reliable and unreliable cells across the borders of non-overlapping or transient regions.} The local-correlation radius \rev{$r_k$} is progressively reduced from coarse to fine levels, enabling broad motion tolerance at low resolution and more precise geometric correction at high resolution.

After Refiner2 at \(\ell_2=1/4\), the residual and matchability predictions are propagated to the \(1/2\) and \(1/1\) grids by bilinear and nearest-neighbor interpolation, respectively, without additional learnable refinement blocks. This design avoids expensive high-resolution convolutions, while keeping the registration branch efficient and stable.

Finally, let \(\Delta W_{1/1}^{\mathcal A}\) and \(M_{1/1}^{\mathcal A}\) denote the full-resolution residual displacement field and matchability map in the affine-aligned frame. For a normalized reference coordinate \(p\), with homogeneous representation \(\tilde p=[p^\top,1]^\top\), its affine-aligned coordinate is
\begin{equation}
    p^{\mathcal A}
    =
    \mathcal A^\star\tilde p.
\end{equation}

\rev{The fine warp composes this affine mapping with the residual bilinearly sampled at $p^{\mathcal A}$, while the fine matchability map is sampled there by nearest neighbor:}
\begin{equation}
    \begin{aligned}
        W_f(p)\equiv W_{1/1}(p)
         & =
        p^{\mathcal A}
        +\Delta W_{1/1}^{\mathcal A}(p^{\mathcal A}), \\
        M_f(p)\equiv M_{1/1}(p)
         & =
        M_{1/1}^{\mathcal A}(p^{\mathcal A}).
    \end{aligned}
\end{equation}

Therefore, LoRetta's final output can be interpreted as the composition of a global affine prior and a learned residual displacement field: the affine prior performs macroscopic geographic alignment, while the residual field is sampled in the affine-aligned frame to compensate for local non-linear distortions caused by topography, viewpoint variation, and imaging perturbations.

\subsection{Matchability Supervision and Training Objectives}
\label{sec:matchability_supervision}

\begin{figure}[t]
    \centering
    \includegraphics[width=\linewidth]{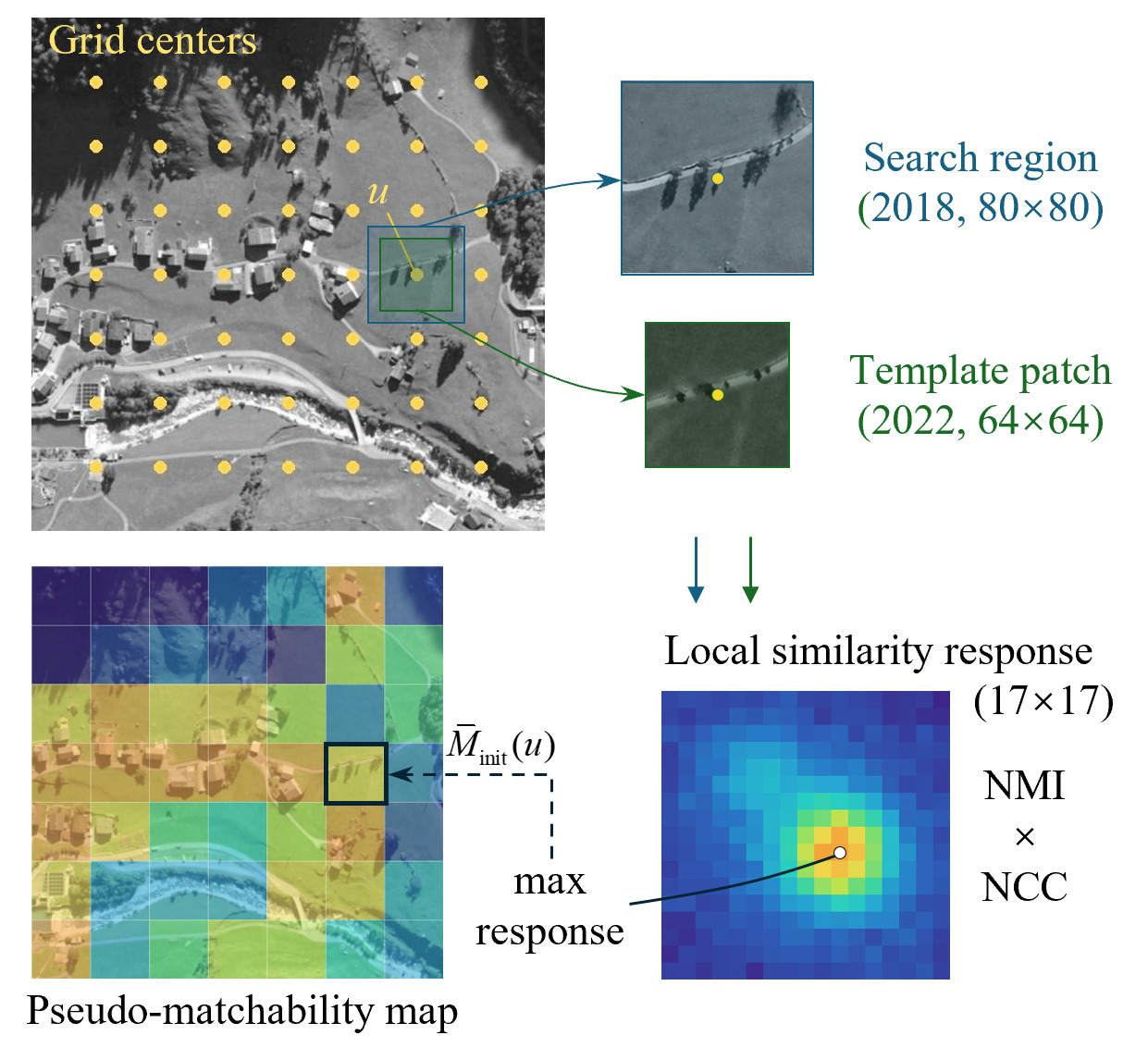}
    \caption{
        \textbf{Pseudo-matchability generation.} A template patch around each coarse grid center is matched within a local search region in the paired image. Normalized mutual information (NMI) and normalized cross-correlation (NCC) response maps are combined and geometrically verified to produce the pseudo-matchability map.
    }
    \label{fig:matchability}
\end{figure}

Before defining the losses, we first specify the pseudo-matchability target used for supervision. Matchability is treated as a data-defined property of a location: it indicates whether a reliable correspondence exists in the paired observation, considering non-overlap, occlusion, seasonal change, and view-dependent parallax. This target is distinct from predictive confidence, which also depends on model capacity, training data, and calibration.

As illustrated in Fig.~\ref{fig:matchability}, pseudo-matchability is generated from classical matching cues rather than manual dense annotations. For each coarse grid location \(u\), we crop a \(64\times64\) template patch and slide it within an \(80\times80\) search region in the paired image, producing \(17\times17\) NMI~\cite{studholme1999nmi} and NCC~\cite{lewis1995ncc} response maps over the displacement domain \(\mathcal D=\{-8,\ldots,8\}^2\). After normalizing both responses to \([0,1]\), we define their hybrid response for each \(\delta\in\mathcal D\) as
\begin{equation}
    H_u(\delta)
    =
    \operatorname{NMI}_u(\delta)
    \operatorname{NCC}_u(\delta).
\end{equation}

Let \(\delta_u^{\mathrm{NMI}}\), \(\delta_u^{\mathrm{NCC}}\), and \(\delta_u^H\) denote the peak displacements of the NMI, NCC, and hybrid responses, respectively:
\begin{equation}
    \begin{aligned}
        \delta_u^{\mathrm{NMI}}
         & =\argmax_{\delta\in\mathcal D}
        \operatorname{NMI}_u(\delta),     \\
        \delta_u^{\mathrm{NCC}}
         & =\argmax_{\delta\in\mathcal D}
        \operatorname{NCC}_u(\delta),     \\
        \delta_u^H
         & =\argmax_{\delta\in\mathcal D}
        H_u(\delta).
    \end{aligned}
\end{equation}
A coarse location is considered verified matchable only when the hybrid peak is sufficiently strong, the NMI and NCC peaks agree spatially, and the hybrid peak is not located on the search boundary. With \(S_u=H_u(\delta_u^H)\), we define the binary verified-matchability indicator as
\begin{equation}
    G_u =
    \mathbf{1}\!\left[
        S_u \ge 0.05
        \land
        \left\|\delta_u^{\mathrm{NMI}}
        -\delta_u^{\mathrm{NCC}}\right\|_2 \le 2
        \land
        \delta_u^{H}\notin\partial\mathcal D
        \right],
\end{equation}
where \(\partial\mathcal D\) denotes the boundary of the displacement domain.

This verification is applied in a three-stage coarse-to-fine process. The first two stages estimate intermediate thin-plate spline (TPS) warps from verified cells to reduce the residual search range, while the final stage produces \(G_u\). The initial low-resolution pseudo-matchability map is defined as
\begin{equation}
    \bar M_{\mathrm{init}}(u)
    =
    \begin{cases}
        1,                               & G_u=1, \\
        \min\!\left(S_u/0.05,0.5\right), & G_u=0.
    \end{cases}
\end{equation}
Verified locations therefore receive full matchability, whereas unverified locations retain a capped soft score.

The initial map \(\bar M_{\mathrm{init}}\) is then propagated through the same geometric transformations used to construct the training pair: it is upsampled with nearest-neighbor interpolation, warped by the synthesized residual perturbation when local deformation is applied, cropped and resized with the image pair, and set to zero wherever the corresponding warp-field target leaves the valid sensed-image crop. We denote the resulting full-resolution pseudo-matchability target by \(\bar M_{1/1}\).

The same data generation process provides the dense warp-field target. The sampled affine crop transform, optionally composed with the synthesized residual perturbation, maps each reference pixel to its target coordinate in the sensed image, yielding the full-resolution target \(\bar W_{1/1}\).

For multi-scale supervision, \(\bar W_{1/1}\) is bilinearly resized and \(\bar M_{1/1}\) is nearest-neighbor resized to each prediction grid. Since the warp field stores normalized reference-to-sensed target coordinates, the coordinate values remain in the same normalized coordinate system across scales; the resulting targets are denoted by \(\bar W_\ell\) and \(\bar M_\ell\).

With these targets, LoRetta is trained according to the localization-and-registration decomposition. Let \(\mathcal S=\{1/16,1/8,1/4,1/2,1/1\}\) denote the supervised registration scales. For each supervised scale \(\ell\in\mathcal S\), the affine-frame predictions are composed as
\begin{equation}
    W_\ell(p)
    =
    p^{\mathcal{A}}
    +
    \Delta W_\ell^{\mathcal{A}}(p^{\mathcal{A}}),
    \qquad
    M_\ell(p)
    =
    M_\ell^{\mathcal{A}}(p^{\mathcal{A}}),
\end{equation}
where \(p^{\mathcal{A}}=\mathcal{A}^\star\tilde p\) as defined above.

For geometric regression, we supervise the fully matchable locations
\begin{equation}
    \Omega_\ell
    =
    \{p\mid \bar M_\ell(p)=1\}.
\end{equation}
At each scale \(\ell\in\mathcal S\), we average the robust coordinate error over \(\Omega_\ell\), using a small \(\epsilon>0\) to avoid division by zero:
\begin{equation}
    \mathcal L_{\mathrm{geo}}^\ell
    =
    \frac{1}{|\Omega_\ell|+\epsilon}
    \sum_{p\in\Omega_\ell}
    \rho\!\left(W_\ell(p)-\bar W_\ell(p)\right).
\end{equation}
To reduce sensitivity to noisy pseudo-warp targets, we use the Cauchy/Lorentzian form of the Barron robust loss~\cite{barron2019robustloss}. For the coordinate residual \(r=(r_1,r_2)^\top=W_\ell(p)-\bar W_\ell(p)\) and robust scale \(c>0\), the penalty is
\begin{equation}
    \rho(r)
    =
    \sum_{d=1}^{2}
    \log\!\left(
    1+\frac{1}{2}
    \left(\frac{r_d}{c}\right)^2
    \right),
\end{equation}
\rev{where $r_d$ is the horizontal ($d=1$) or vertical ($d=2$) error.}

While geometric regression uses the hard fully matchable set \(\Omega_\ell\), matchability prediction is supervised by the full soft target \(\bar M_\ell\) using class-balanced binary cross-entropy (BCE):
\begin{equation}
    \mathcal L_{\mathrm{mat}}^\ell
    =
    \operatorname{BCE}\!\left(
    \operatorname{MaxPool}_{k_\ell}(M_\ell),
    \bar M_\ell
    \right).
\end{equation}
Here \(\operatorname{BCE}(\cdot,\cdot)\) is evaluated in probability space; \(M_\ell\in[0,1]\) is the predicted matchability probability map, and \(\bar M_\ell\in[0,1]\) is the soft pseudo-matchability target. Since the pseudo-matchability target is generated from coarse template centers and then propagated to dense grids, exact pixel-level alignment between the target and the predicted matchability peak would be unnecessarily strict. We therefore apply max-pooling to \(M_\ell\) before the BCE loss, allowing small shifts caused by label quantization, residual perturbations, and multi-scale resizing while still discouraging high matchability responses around non-matchable regions. The kernel size \(k_\ell\) is scale-dependent: for \(\ell\in\mathcal S\), \(k_\ell=(1,3,5,9,17)\) in the order of \(\mathcal S\). These values keep the tolerance region roughly comparable in original image coordinates.

The coarse localization branch is additionally supervised by a coarse matching classification loss. Let
\(\bar m_i=\bar M_{1/1}(p_i)\) denote the pseudo-matchability target at the center \(p_i\) of reference token \(i\). We supervise only tokens with binary targets:
\begin{equation}
    \Omega_c
    =
    \{i\mid \bar m_i\in\{0,1\}\}.
\end{equation}
For an in-frame target with \(\bar m_i=1\), \(\mathcal N_i\) contains its four neighboring sensed-grid tokens, with bilinear weights \(\omega_{ij}\geq0\) satisfying \(\sum_{j\in\mathcal N_i}\omega_{ij}=1\); an out-of-frame target with \(\bar m_i=0\) is assigned to the no-match bin \(\varnothing\). The loss is
\begin{equation}
    \begin{aligned}
        \mathcal L_{\mathrm{cls}}
        =
        -\frac{1}{|\Omega_c|+\epsilon}
        \sum_{i\in\Omega_c}
        \Bigg(
         & \bar m_i
        \sum_{j\in\mathcal N_i}
        \omega_{ij}\log P_{ij}
                            {}+ \\
         & (1-\bar m_i)
        \log P_{i,\varnothing}
        \Bigg).
    \end{aligned}
\end{equation}
This loss aligns the coarse matching distribution with the dense warp-field target before residual registration.

We further stabilize the global localization branch with affine supervision. The affine transform used in data construction provides a target matrix \(\bar{\mathcal{A}}\in\mathbb{R}^{2\times3}\), following the same coordinate convention as the predicted affine prior \(\mathcal{A}^\star\). We supervise the predicted affine matrix using
the squared Frobenius error averaged over its six entries:
\begin{equation}
    \mathcal L_{\mathrm{aff}}
    =
    \frac{1}{6}
    \left\|
    \mathcal{A}^\star-\bar{\mathcal{A}}
    \right\|_F^2.
\end{equation}
This supervision encourages the localization branch to recover the affine geometry used to initialize dense registration.

The final training objective is
\begin{equation}
    \mathcal L
    =
    \sum_{\ell\in\mathcal S}\alpha_\ell
    \left(
    \mathcal L_{\mathrm{geo}}^\ell
    + \lambda_{\mathrm{mat}}\mathcal L_{\mathrm{mat}}^\ell
    \right)
    + \lambda_{\mathrm{cls}}\mathcal L_{\mathrm{cls}}
    + \lambda_{\mathrm{aff}}\mathcal L_{\mathrm{aff}}.
\end{equation}
The scale weights \(\alpha_\ell\) emphasize later, higher-resolution registration outputs, while \(\lambda_{\mathrm{mat}}\), \(\lambda_{\mathrm{cls}}\), and \(\lambda_{\mathrm{aff}}\) balance matchability supervision, coarse matching classification, and affine supervision. This objective trains LoRetta to estimate dense correspondence fields, predict where those correspondences are reliable, and maintain an affine initialization consistent with the global geometry of the training pair.

\section{DATASET}
\label{sec:dataset}

\begin{figure*}[!t]
    \centering
    \includegraphics[width=1.0\linewidth]{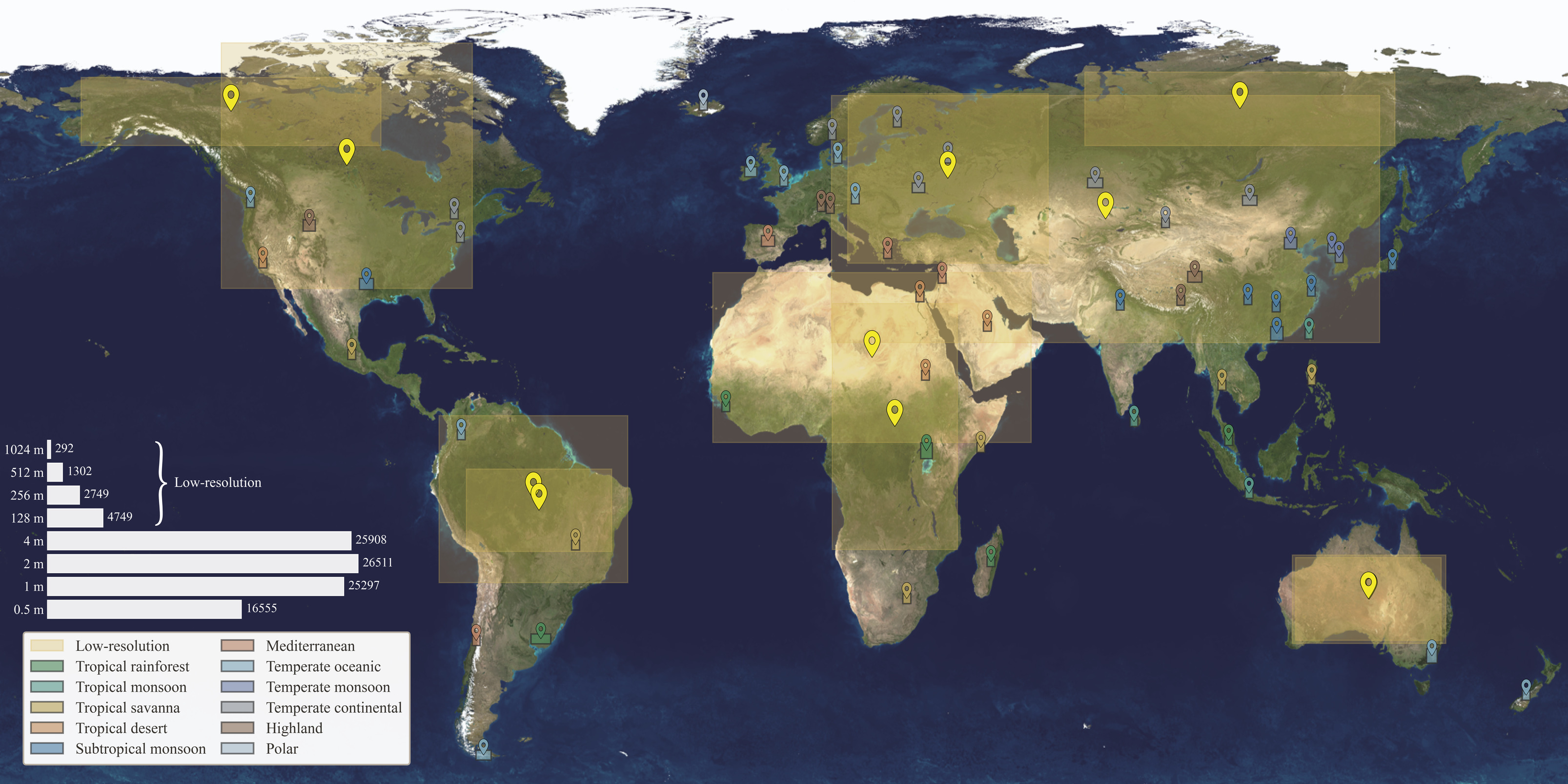}
    \caption{
        \textbf{Global geographic distribution and resolution composition of LEVIR-GM.} Colored polygons and markers show high-resolution regions grouped by climate type, while yellow translucent polygons denote low-resolution coverage areas. \protect\rev{Each white bar gives the number of aligned base-layer image pairs at one spatial resolution; bars from 128 m to 1024 m form the low-resolution group, and all sum to 103,363 pairs before augmentation.}
    }
    \label{fig:dataset_distribution}
\end{figure*}

\subsection{LEVIR-GM Overview}

To address the data gap in global-scale dense matching discussed in Section~\ref{sec:introduction}, we construct LEVIR-GM, a large-scale optical remote sensing matching dataset built from multi-temporal observations collected between 2018 and 2022. Unlike conventional remote sensing registration datasets that are usually restricted to a small number of cities, a narrow spatial-resolution range, or a fixed perturbation protocol, LEVIR-GM is designed to expose dense matchers to three coupled challenges: long-term visual change, cross-resolution observation, and complex geometric displacement.

LEVIR-GM follows a two-layer construction. The aligned base layer contains 103,363 high-quality multi-temporal image pairs. Each aligned pair has a size of \(1280\times1280\), with spatial resolutions ranging from 0.5 m to 1024 m. The augmented layer is generated from this aligned base and contains 826,904 image pairs. Each augmented pair is saved at \(512\times512\), with an effective resolution range of 1.0--2048 m. Thus, the aligned layer provides realistic multi-temporal visual content, while the augmented layer provides controlled geometric perturbations and dense supervision for training and evaluation.

As shown in Fig.~\ref{fig:dataset_distribution}, the aligned base layer covers 67 geographic regions across six continents, including 56 high-resolution regions and 11 large low-resolution coverage areas. The high-resolution regions are distributed over Asia, Europe, Africa, North America, South America, and Oceania, with 23, 12, 7, 7, 5, and 2 regions, respectively. To reduce geographic and climatic bias, these regions cover 11 representative climate types, including tropical rainforest, tropical monsoon, tropical savanna, tropical desert, subtropical monsoon, Mediterranean, temperate oceanic, temperate monsoon, temperate continental, highland, and polar climates. Fig.~\ref{fig:dataset_distribution} also summarizes the aligned-layer resolution composition from 0.5 m to 1024 m, while Table~\ref{tab:dataset_benchmark_compare} separates the aligned and augmented layers to make their different roles explicit. This global geographic and multi-resolution coverage makes LEVIR-GM suitable for evaluating whether a matcher can generalize beyond local scene statistics.

\begin{table*}[!t]
    \caption{
        Comparison with representative remote sensing image matching datasets
    }
    \label{tab:dataset_benchmark_compare}
    \centering
    \scriptsize
    \setlength{\tabcolsep}{0.45pt}
    \renewcommand{\arraystretch}{1.05}
    \begin{tabular*}{0.76\textwidth}{@{\extracolsep{\fill}}C{0.17\textwidth}C{0.11\textwidth}C{0.065\textwidth}C{0.08\textwidth}C{0.115\textwidth}C{0.095\textwidth}C{0.08\textwidth}@{}}
        \toprule
        Dataset                                  & Modality        & Scale & \shortstack{Image Size\\(px)} & \shortstack{Spatial Resolution\\(m/px)} & Temporal Coverage & Matchability   \\
        \midrule

        DeepAerialMatching~\cite{park2020twostream} & optical-optical & 9K & 240        & -                                           & 2015/2017/2019    & \(\times\)     \\

        MID~\cite{feng2022mid}                   & optical-optical & 4.1K  & 512        & 0.8                                         & -                 & \(\times\)     \\

        RSI Matching Dataset~\cite{li2021semantictemplate} & optical-optical / optical-SAR          & 1K    & 512        & 0.23/3.75/30                                & -                 & \(\times\)     \\

        SOPatch~\cite{xu2023sopatch}             & optical-SAR     & 666K  & 64         & 1--10                                       & -                 & \(\times\)     \\

        QXS-SAROPT~\cite{huang2021qxs}           & optical-SAR     & 20K   & 256        & 1                                           & -                 & \(\times\)     \\

        SEN1-2~\cite{schmitt2018sen12}           & optical-SAR     & 282K  & 256        & 10                                          & 2016--2017        & \(\times\)     \\

        SARptical~\cite{wang2018sarptical}       & optical-SAR     & 90K   & 112        & 0.2--1                                      & 2009--2013        & \(\times\)     \\

        OSEval~\cite{xiang2023globaltolocal}     & optical-SAR     & 1.2K  & 1200--5500 & 0.3--0.56                                   & -                 & \(\times\)     \\

        OS-dataset~\cite{xiang2020osdataset}     & optical-SAR     & 11K   & 256/512    & 1                                           & -                 & \(\times\)     \\

        SpaceNet 6~\cite{shermeyer2020spacenet6} & optical-SAR     & 3K    & 900        & 0.5--2                                      & 2019              & \(\times\)     \\

        3MOS~\cite{ye20253mos}                   & optical-SAR     & 113K  & 256        & 3.5--12.5                                   & 2012--2024        & \(\times\)     \\

        SOMA-1M~\cite{wu2026soma1m}              & optical-SAR     & 1.3M  & 512        & 0.5--10                                     & -                 & \(\times\)     \\
        \midrule
        LEVIR-GM (aligned)                       & optical-optical & 103K  & 1280       & 0.5--1024                                   & 2018--2022        & \(\checkmark\) \\
        LEVIR-GM (augmented)                     & optical-optical & 827K  & 512        & 1--2048                                     & 2018--2022        & \(\checkmark\) \\
        \bottomrule
    \end{tabular*}
\end{table*}

\begin{table}[!t]
    \caption{
        Comparison of training transformation protocols for end-to-end regression methods in remote sensing image matching
    }
    \label{tab:dataset_transformation_compare}
    \centering
    \scriptsize
    \setlength{\tabcolsep}{0.45pt}
    \renewcommand{\arraystretch}{1.05}
    \begin{tabular*}{\columnwidth}{@{\extracolsep{\fill}}C{0.27\columnwidth}C{0.18\columnwidth}C{0.12\columnwidth}C{0.12\columnwidth}C{0.12\columnwidth}C{0.12\columnwidth}@{}}
        \toprule
        Method                                            & \shortstack{Transform\\Type} & Rotation           & Scale               & Shear        & \shortstack{Local\\Nonrigid} \\
        \midrule
        Park~\cite{park2020twostream}                     & affine         & -                  & -                   & -            & \(\times\)      \\
        Huang~\cite{huang2021videosar}                    & rigid          & -                  & -                   & -            & \(\checkmark\)  \\
        Papadomanolaki~\cite{papadomanolaki2021multistep} & affine         & -                  & -                   & -            & \(\checkmark\)  \\
        MU-Net~\cite{ye2022munet}                         & affine         & \(\pm180^{\circ}\) & 0.5--2.0            & \(\pm\pi/6\) & \(\times\)      \\
        MID~\cite{feng2022mid}                            & affine         & -                  & -                   & -            & \(\checkmark\)  \\
        Li~\cite{li2023fusionregistration}                & homography     & -                  & -                   & -            & \(\times\)      \\
        GcrDfNet~\cite{chang2023gcrdfnet}                 & affine         & \(\pm30^{\circ}\)  & 0.8--1.2            & -            & \(\times\)      \\
        OSFlowNet~\cite{zhang2023osflownet}               & affine         & -                  & -                   & -            & \(\checkmark\)  \\
        OS\(^3\)Flow~\cite{sun2024os3flow}                     & affine         & \(\pm10^{\circ}\)  & 0.9--1.1            & -            & \(\checkmark\)  \\
        \revrow GDROS~\cite{sun2025gdros}                       & affine         & \(\pm20^{\circ}\)  & 0.8--1.2            & -            & \(\times\)      \\
        \midrule
        LEVIR-GM                                          & affine         & \(\pm180^{\circ}\) & 0.8--4.0            & \(\pm\pi/6\) & \(\checkmark\)  \\
        \bottomrule
    \end{tabular*}
\end{table}

\subsection{Augmented Layer and Matchability Supervision}

Starting from the aligned base layer described above, each augmented sample is generated by applying controlled geometric perturbations to a real aligned pair and by deriving the corresponding geometric and matchability supervision. The generator first samples an affine transform consisting of scale, translation, rotation, and shear, checks whether the transformed crop remains valid, and then produces a dense warp-field target. The default crop size is \(1024\times1024\), and the final sample is resized to \(512\times512\). Unlike transformation pipelines that warp an image into a fixed canvas and retain padded regions, LEVIR-GM uses validity checking and crop selection to avoid retaining artificial black-border regions in the released image pairs. This removes a low-level artifact that could bias a model toward dataset-specific padding patterns, degrade training stability, and reduce cross-dataset generalization.

The perturbation range is resolution-aware. The main augmented layer is generated from high-resolution aligned pairs at 0.5 m, 1 m, 2 m, and 4 m, for which the scale factor is sampled from 1.0 to 4.0, with a full \(\pm\pi\) rotation range and a shear range of \(\pm\pi/6\). Hierarchical local non-rigid perturbations are further added to simulate residual local distortions. The perturbation strength depends on the source resolution: the numbers of local perturbation centers are 8, 16, 24, and 32 for 0.5 m, 1 m, 2 m, and 4 m, respectively, while the corresponding maximum amplitudes are 32, 24, 16, and 8 pixels. In addition, a small supplementary set is generated from low-resolution aligned pairs at 128 m, 256 m, 512 m, and 1024 m \rev{to cover the appearance of coarse imagery.} For these low-resolution pairs, the scale factor is sampled from 0.8 to 1.5, with the same rotation and shear ranges, and no local non-rigid perturbation is added. This design reflects the intended decomposition of global-scale matching: a global affine prior handles the dominant macro-alignment, \rev{including scale mismatches between sensors~\cite{yan2016landsat} or viewing geometries,} while residual dense registration compensates for local distortions when fine local structures are available.

The matchability supervision in the augmented layer reuses the pseudo-matchability generation process introduced in Section~\ref{sec:matchability_supervision}. For each base pair, the template matching stage produces hybrid NMI/NCC responses and a verified good-mask. Verified reliable cells are assigned a matchability value of one, while unverified locations retain only a capped soft score to avoid treating uncertain responses as fully matchable. During augmentation, the certainty map and good-mask are transformed consistently with the image pair, cropped, resized, and set to zero wherever the warp-field target falls outside the valid sensed crop. Consequently, LEVIR-GM provides not only image pairs and geometric labels, but also dataset-native pseudo-matchability for learning when dense warp-field predictions should be trusted or rejected.

\subsection{Comparison with Existing Datasets and Protocols}

Tables~\ref{tab:dataset_benchmark_compare} and~\ref{tab:dataset_transformation_compare} compare LEVIR-GM with existing remote sensing matching datasets and training protocols, where ``-'' denotes settings that are not explicitly reported in the original paper or public material. At the dataset level, LEVIR-GM is distinguished by its scale, global geographic coverage, long temporal span, and broad spatial-resolution range. Most existing benchmarks focus on high- or medium-resolution local scenes, optical-SAR alignment, or task-specific registration settings, whereas LEVIR-GM \rev{covers two resolution regimes: 0.5--4 m per pixel, as used in change analysis and UAV localization, and 128--1024 m per pixel, where texture becomes weak and large-scale geographic semantics dominate, as in the downsampled Sentinel-2 basemaps used for astronaut photograph localization~\cite{berton2024earthloc}}. Another key difference is matchability supervision. In natural-image matching, valid-match confidence can often be derived from camera poses, depth maps, and reprojection errors. Such supervision is rarely available in multi-temporal remote sensing because dense depth, accurate camera geometry, and stable pixel-level correspondences are difficult to obtain, while clouds, shadows, seasonal changes, land-cover changes, parallax, and non-overlap make many regions inherently unreliable for dense matching. LEVIR-GM therefore provides dataset-native pseudo-matchability together with dense warp-field labels, which can support not only LoRetta but also the training of sparse, semi-dense, and dense matching methods by supervising keypoint reliability, coarse correspondence confidence, or dense matchability/uncertainty maps.

At the protocol level, existing end-to-end remote sensing registration methods commonly rely on synthetic affine, homography, rigid, or local-flow perturbations, but their exact training ranges are often incompletely reported or restricted to relatively mild geometric changes. LEVIR-GM makes the augmentation protocol explicit and reproducible. It covers full-angle rotation, large-scale variation, affine shear, and resolution-aware local non-rigid perturbations, while preserving the real multi-temporal visual variation inherited from the aligned base pairs. In addition, the augmented layer is generated with validity checking and crop selection, avoiding artificial black borders caused by geometric transformations. As a result, LEVIR-GM is not merely a larger collection of paired images, but a controlled training and evaluation testbed for decomposing global-scale matching into coarse localization and residual dense registration: the affine component tests macroscopic alignment, the local perturbations test dense residual registration, and the pseudo-matchability \rev{labels} test whether unreliable regions can be rejected during learning and evaluation.

\section{EXPERIMENTS}
\label{sec:experiments}

\subsection{Experimental Setup}

The controlled benchmark experiments are conducted on LEVIR-GM described in Section~\ref{sec:dataset}. The augmented layer is split into training, validation, and test sets with a ratio of \(80\%/10\%/10\%\), and the main LEVIR-GM results are reported on the held-out test split. For this benchmark, each image pair is evaluated at \(512\times512\) resolution, matching the resolution used during training. Pixel coordinates and error thresholds are measured in this resized image frame.

We compare LoRetta with representative sparse, semi-dense, and dense matchers, including SIFT, SuperPoint with LightGlue (SP+LG), LoFTR, DKM, RoMa, and RoMa v2. All methods take the same resized image pairs as input. For fair comparison across methods with different output densities, we evaluate every method through the same registration protocol. Each matcher first provides a set of confident correspondences or dense correspondence candidates. These correspondences are filtered by random sample consensus (RANSAC)~\cite{fischler1981ransac}, and a thin-plate spline (TPS) transform~\cite{bookstein1989tps} is then fitted from the remaining inliers. The fitted TPS transform is rasterized on the \(512\times512\) reference grid to obtain the evaluated dense reference-to-sensed warp field, denoted by \(\hat W\). Thus sparse, semi-dense, and dense matchers are all scored as dense registration systems under the same post-processing protocol.

For learned baselines, SP+LG, LoFTR, DKM, and RoMa are initialized from GIM weights and fine-tuned on LEVIR-GM. We use GIM initialization because GIM provides strong cross-view matching priors learned from large-scale natural-image matching, as discussed in Section~\ref{sec:training_pipelines}, and therefore gives these baselines a competitive starting point before remote sensing adaptation. RoMa v2 is fine-tuned from its official released weights, while SIFT is used without learning-based fine-tuning. All learned baselines are fine-tuned following their official training recipes when available, and the best checkpoint of each method is selected on the validation split under the same AUC metric.

To isolate the effect of the localization-and-registration design from the visual backbone, LoRetta uses the same frozen DINOv3~\cite{simeoni2025dinov3} ViT-L architecture and weights as RoMa v2. The DINOv3 feature extractor provides \(1/16\)-resolution descriptors from layer indices 11 and 17. The resulting \(32\times32\) token grid is processed by the multi-view Transformer and match-embedding decoder, producing the coarse warp field and matchability map used by the affine locator. The affine locator uses \(\tau_A=0.3\) to select reliable coarse correspondences for global affine fitting. The localization-guided registration branch uses a trainable VGG19~\cite{simonyan2015vgg} feature pyramid. The local-correlation radii \rev{$r_k$} of the three learnable refiners at \(1/16\), \(1/8\), and \(1/4\) resolution are \(5\), \(3\), and \(2\), respectively; the \(1/2\) and \(1/1\) outputs are obtained by interpolation.

LoRetta is implemented in PyTorch 2.6.0 with PyTorch Lightning 2.5.0 and trained on a single NVIDIA RTX 4090 GPU using AdamW~\cite{loshchilov2019adamw} with weight decay 0.1. Training is conducted in two stages. In the first stage, we train the coarse localization branch with batch size 16 for \(0.83\)M iterations and an initial learning rate of \(1\times10^{-5}\). Only the coarse localization output and the affine prior are supervised in this stage. In the second stage, we freeze the localization branch and train the dense registration branch with batch size 4 for \(3.31\)M iterations and an initial learning rate of \(5\times10^{-6}\). The multi-scale supervision weights \(\alpha_\ell\) are \(0.1,0.1,0.1,0.2,\) and \(0.5\) for \(1/16,1/8,1/4,1/2,\) and \(1/1\) predictions, respectively. The Cauchy/Lorentzian robust loss uses scale \(c=0.1\). We set \(\lambda_{\mathrm{mat}}=0.01\), \(\lambda_{\mathrm{cls}}=10^{-4}\), and \(\lambda_{\mathrm{aff}}=0.05\), with the geometric regression term kept at unit weight.

For evaluation, both the predicted dense warp \(\hat W\) and the target warp \(\bar W_{1/1}\) are represented in the \(512\times512\) pixel coordinate frame. Let \(e(p)=\|\hat W(p)-\bar W_{1/1}(p)\|_2\) be the endpoint error in pixels at reference pixel \(p\). Following the matchability supervision defined in Section~\ref{sec:matchability_supervision}, the metrics are computed only on fully matchable pixels:
\begin{equation}
    \Omega
    =
    \{p \mid \bar M_{1/1}(p)=1\}.
\end{equation}
Percentage of Correct Keypoints (PCK) at threshold \(t\) is computed as
\begin{equation}
    \operatorname{PCK}(t)
    =
    \frac{1}{|\Omega|}
    \sum_{p\in\Omega}\mathbf{1}\!\left[e(p)\leq t\right].
\end{equation}
We report PCK at \(1,2,3,5,\) and \(10\) pixels. Area Under Curve (AUC) is computed as the normalized trapezoidal area under the PCK curve over the threshold set \(\mathcal T=\{0.5,1,1.5,2,2.5\}\cup\{3,4,\ldots,10\}\) pixels. Let \(\mathcal T=\{t_i\}_{i=1}^{K}\) in ascending order. We compute
\begin{equation}
    \operatorname{AUC}
    =
    \frac{1}{t_K-t_1}
    \sum_{i=1}^{K-1}
    \frac{t_{i+1}-t_i}{2}
    \left[
        \operatorname{PCK}(t_i)
        +\operatorname{PCK}(t_{i+1})
        \right].
\end{equation}
Inference time is measured for matcher inference only, excluding data loading and the unified RANSAC/TPS scoring layer, and is averaged per \(512\times512\) image pair.

\subsection{Main Results on LEVIR-GM}

\begin{figure}[!t]
    \centering
    \includegraphics[width=\linewidth]{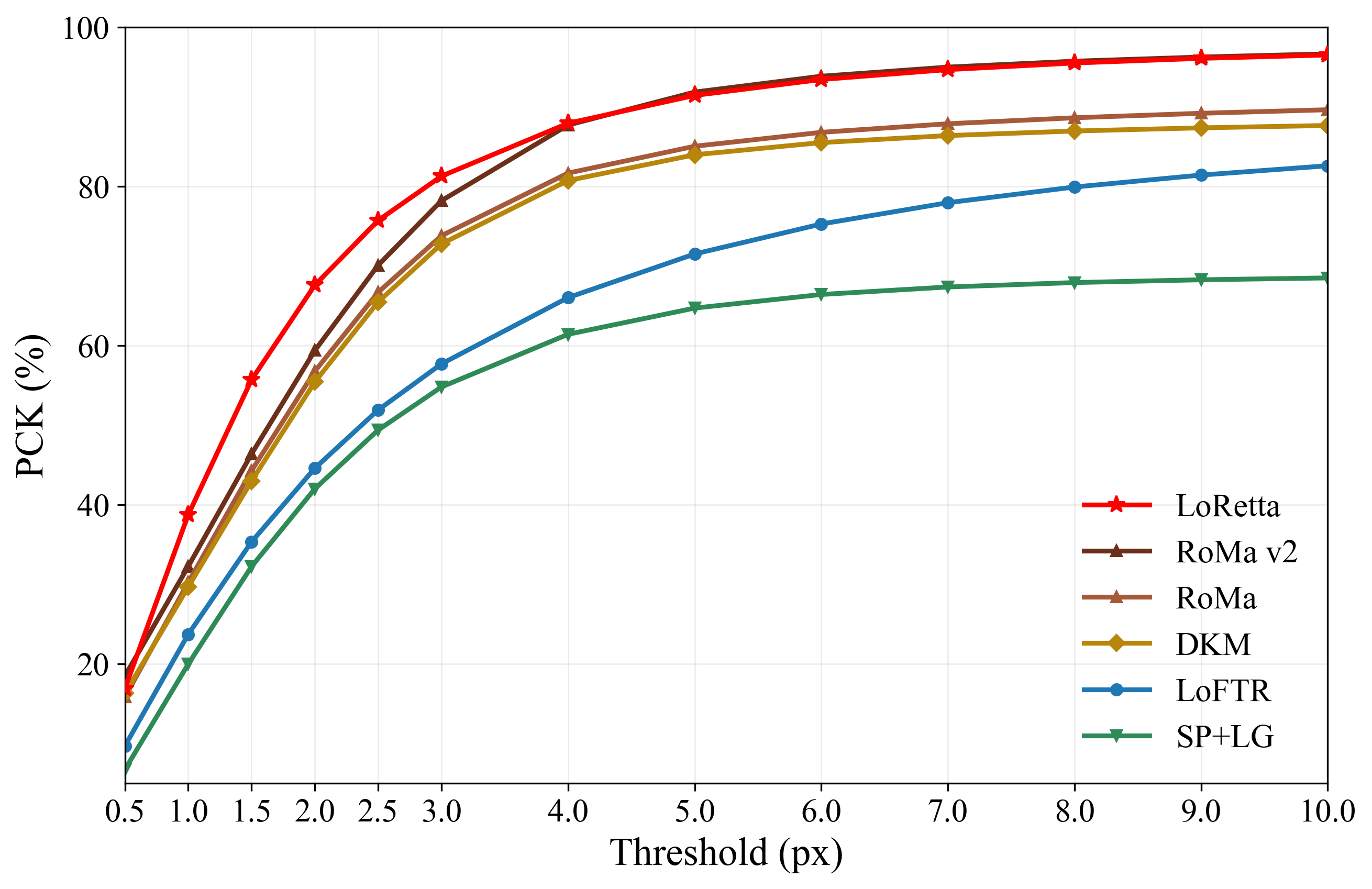}
    \caption{
        \textbf{PCK curves on LEVIR-GM.} The curves compare LoRetta with all learned baselines under pixel-error thresholds from 0.5 to 10 pixels.
    }
    \label{fig:exp_pck_curves}
\end{figure}

Table~\ref{tab:main_comparison} and Fig.~\ref{fig:exp_pck_curves} summarize the overall comparison. LoRetta achieves the best AUC of \(83.3\%\), improving over the strongest baseline RoMa v2 by \(1.6\) points. The gain is concentrated at strict pixel-error thresholds: compared with RoMa v2, LoRetta improves PCK by \(6.5\), \(8.2\), and \(3.1\) points at 1, 2, and 3 pixels, respectively, reaching \(38.7\%\), \(67.6\%\), and \(81.3\%\).

At relaxed thresholds, RoMa v2 is marginally higher at 5 and 10 pixels, where the benchmark becomes close to saturation for modern dense matchers. However, RoMa v2 requires \(124.1\) ms per pair, while LoRetta takes \(64.8\) ms, making LoRetta about \(1.9\times\) faster. Sparse and semi-dense methods are faster in some cases, but their AUC is notably lower because their correspondences lead to less accurate dense alignment after the unified TPS fitting, especially in weak-texture or heavily transformed remote sensing scenes.

\begin{table}[!t]
    \centering
    \caption{
        Quantitative comparison of sparse, semi-dense, and dense methods on LEVIR-GM using PCK, AUC, and inference time. PCK thresholds are measured in pixels
    }
    \label{tab:main_comparison}
    \scriptsize
    \setlength{\tabcolsep}{0.45pt}
    \renewcommand{\arraystretch}{1.05}
    \begin{tabular*}{\columnwidth}{@{\extracolsep{\fill}}C{0.14\columnwidth}C{0.18\columnwidth}*{7}{C{0.085\columnwidth}}@{}}
        \toprule
        \multirow{2}{*}{\raisebox{-0.55ex}{Type}} & \multirow{2}{*}{\raisebox{-0.55ex}{Method}}                                 & \multicolumn{5}{c}{PCK (\%) \(\uparrow\)}
        & \multirow{2}{*}{\raisebox{-3.0ex}{\shortstack{AUC\\(\%) \(\uparrow\)}}}
        & \multirow{2}{*}{\raisebox{-3.0ex}{\shortstack{Time\\(ms) \(\downarrow\)}}}                                                                                                                                                               \\
        \cmidrule(lr){3-7}
        &                                                         & 1 px                                   & 2 px             & 3 px             & 5 px             & 10 px            &                  &                  \\
        \midrule
        \multirow{3}{0.14\columnwidth}{\centering\shortstack[c]{Sparse/\\Semi-Dense}}
        & SIFT                                                    & 12.6                                      & 19.2             & 22.5             & 25.5             & 27.7             & 23.6             & 166.0            \\
        & SP+LG                                                   & 20.0                                      & 42.0             & 54.8             & 64.7             & 68.5             & 57.3             & \textbf{24.0}    \\
        & LoFTR                                                   & 23.7                                      & 44.6             & 57.7             & 71.5             & 82.6             & 64.9             & \underline{32.1} \\
        \midrule
        \multirow{4}{0.14\columnwidth}{\centering Dense}
        & DKM                                                     & 29.7                                      & 55.4             & 72.8             & 84.0             & 87.6             & 74.7             & 33.4             \\
        & RoMa                                                    & 30.2                                      & 56.8             & 73.8             & 85.0             & 89.6             & 76.0             & 119.6            \\
        & RoMa v2                                                 & \underline{32.2}                          & \underline{59.4} & \underline{78.2} & \textbf{91.8}    & \textbf{96.6}    & \underline{81.7} & 124.1            \\
        & LoRetta (Ours)                                          & \textbf{38.7}                             & \textbf{67.6}    & \textbf{81.3}    & \underline{91.4} & \underline{96.5} & \textbf{83.3}    & 64.8             \\
        \bottomrule
    \end{tabular*}
\end{table}

\begin{figure*}[!t]
    \centering
    \includegraphics[width=\textwidth]{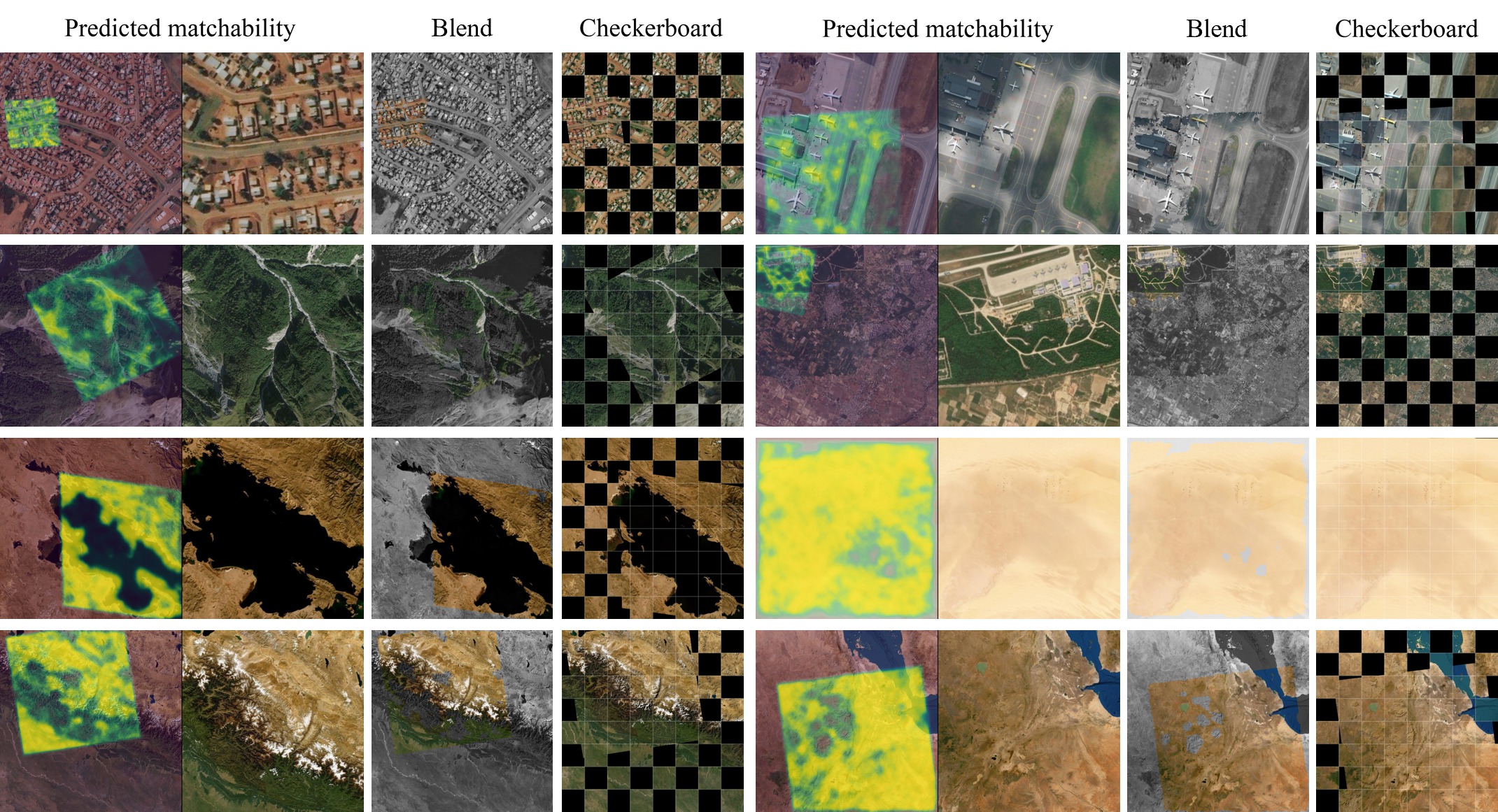}
    \caption{
        \textbf{Qualitative dense matching results of LoRetta on LEVIR-GM.} Matchability visualizes the regions predicted to contain reliable correspondences. Blend directly visualizes LoRetta's raw dense prediction by overlaying the high-matchability regions of the warped sensed image in color on the grayscale reference image. Checkerboard shows the registration obtained after the RANSAC and TPS fitting used in the unified evaluation protocol.
    }
    \label{fig:exp_qualitative_result}
\end{figure*}

Fig.~\ref{fig:exp_qualitative_result} further illustrates both LoRetta's native dense predictions and the registration results obtained under the unified evaluation protocol. The matchability maps show that LoRetta concentrates high-confidence predictions on reliable overlapping regions while suppressing changed, ambiguous, or non-matchable areas. The blend views directly visualize the raw dense warp predicted by LoRetta, showing coherent alignment across urban scenes, airports, mountains, water bodies, deserts, low-texture surfaces, and large geometric changes. The TPS-fitted checkerboard views additionally show the final registration obtained through the same post-processing procedure used for quantitative evaluation.

\subsection{Land-Cover-Wise Analysis}

LEVIR-GM is collected from globally distributed regions rather than from a predefined land-cover taxonomy, so the test set does not directly provide semantic group labels. To make this analysis interpretable, we assign post-hoc scene-level labels to the high-resolution test samples with a lightweight grouping procedure. We first build a balanced reference set of visually unambiguous patches for six dominant groups: building, mountain, farmland, vegetation, water, and barren. A frozen DINOv3 encoder maps both reference patches and test samples to feature maps, and image descriptors are obtained by global average pooling. Each test sample is assigned to the group of its nearest reference example in this descriptor space. These labels are used only to partition the evaluation results in Table~\ref{tab:category_comparison}, and are not used for training, model selection, or scoring.

The first six AUC columns correspond to the high-resolution samples grouped by these scene-level labels. The 128--1024 m/pixel subset is reported separately as low-resolution because each crop covers a much larger area and often contains mixed land-cover content, making a single high-resolution semantic label less meaningful. The test sample counts are \(2{,}955\), \(1{,}739\), \(1{,}689\), \(1{,}373\), \(549\), \(1{,}123\), and \(909\) for building, mountain, farmland, vegetation, water, barren, and low-resolution samples, respectively.

\begin{figure*}[!t]
    \centering
    \includegraphics[width=\textwidth]{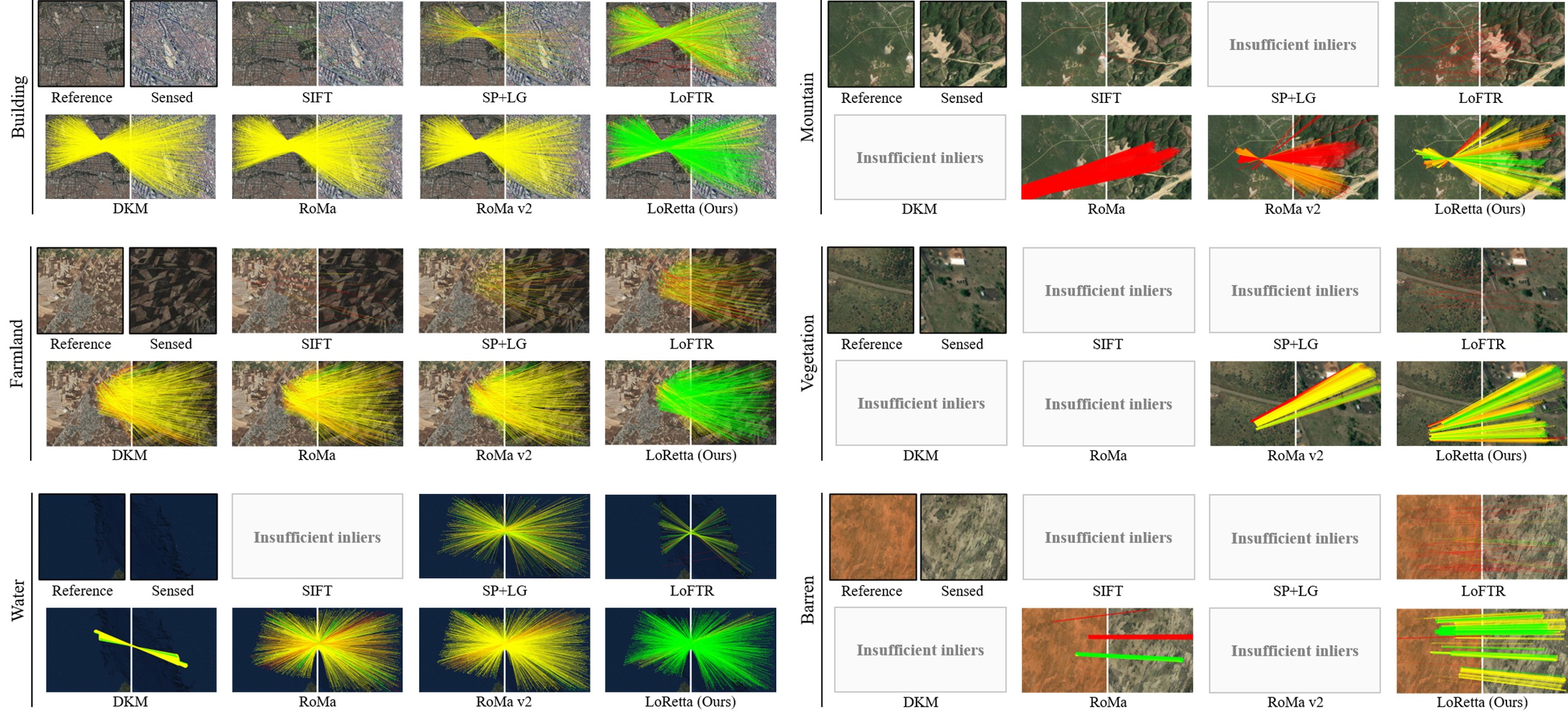}
    \caption{
        \textbf{Qualitative matching results across representative land-cover groups.} The examples cover building, mountain, farmland, vegetation, water, and barren scenes. Each method panel shows spatially sampled correspondences retained after confidence filtering and RANSAC. Correspondence lines are colored by endpoint error \(e\) with respect to the ground-truth dense warp field: green denotes \(e<1\) px, yellow denotes \(1\leq e<3\) px, orange denotes \(3\leq e<5\) px, and red denotes \(e\geq5\) px. A gray crossed panel indicates that fewer than 3 correspondences remain after filtering and therefore cannot provide a reliable visualization.
    }
    \label{fig:exp_category_visualization}
\end{figure*}

\begin{table}[!t]
    \centering
    \caption{
        Land-cover-wise AUC comparison on LEVIR-GM over six post-hoc high-resolution scene groups and the 128--1024 m/pixel low-resolution subset
    }
    \label{tab:category_comparison}
    \scriptsize
    \setlength{\tabcolsep}{0.8pt}
    \renewcommand{\arraystretch}{1.05}
    \begin{tabular*}{\columnwidth}{@{\extracolsep{\fill}}C{0.17\columnwidth}C{0.10\columnwidth}C{0.11\columnwidth}C{0.105\columnwidth}C{0.125\columnwidth}C{0.065\columnwidth}C{0.075\columnwidth}C{0.095\columnwidth}@{}}
        \toprule
        \multirow{2}{*}{\raisebox{-0.55ex}{Method}}                & \multicolumn{7}{c}{AUC (\%) \(\uparrow\)}                                                                                                                                                                   \\
        \cmidrule(l){2-8}
        & Building                      & Mountain & Farmland & Vegetation & Water & Barren & \mbox{Low-res} \\
        \midrule
        SIFT                                   & 22.5                                          & 22.5                     & 12.1                     & 14.0                       & 14.9                  & 13.6                   & 82.6                    \\
        SP+LG                                  & 59.7                                          & 51.9                     & 52.0                     & 49.2                       & 49.4                  & 50.7                   & 95.5                    \\
        LoFTR                                  & 68.5                                          & 61.4                     & 59.0                     & 59.0                       & 58.9                  & 55.8                   & 94.8                    \\
        DKM                                    & 78.1                                          & 70.9                     & 70.4                     & 73.1                       & 69.0                  & 66.5                   & 95.3                    \\
        RoMa                                   & 79.9                                          & 71.6                     & 70.9                     & 70.5                       & 71.2                  & 71.8                   & 97.8                    \\
        RoMa v2                                & 82.9                                          & 79.3                     & 76.6                     & 79.5                       & 76.8                  & 81.1                   & \textbf{98.4}           \\
        LoRetta (Ours)                         & \textbf{84.1}                                 & \textbf{82.3}            & \textbf{78.2}            & \textbf{80.9}              & \textbf{77.6}         & \textbf{83.8}          & 98.3                    \\
        \bottomrule
    \end{tabular*}
\end{table}

Using this partition, Table~\ref{tab:category_comparison} evaluates whether the model remains effective across this semantic diversity rather than only improving the average score. LoRetta obtains the best AUC on all six high-resolution land-cover groups. The largest improvements over RoMa v2 appear in mountain and barren regions, with gains of \(3.0\) and \(2.7\) AUC points, respectively, where strong relief variation, weak texture, and seasonal radiometric changes make methods without explicit localization guidance less stable. The low-resolution subset is close to saturation for strong dense matchers: RoMa v2 is marginally higher by \(0.1\) AUC point, while LoRetta remains essentially tied. This result is consistent with the dataset construction in Section~\ref{sec:dataset}: low-resolution pairs span a broader geographic extent per crop but contain less fine local structure, so they are more dominated by large-scale semantic alignment than by high-resolution local registration.

Fig.~\ref{fig:exp_category_visualization} further provides qualitative comparisons across representative land-cover groups. Under the same visualization and spatial-sampling protocol, LoRetta generally produces more spatially distributed low-error correspondences, whereas several baselines exhibit more correspondences with errors above 3 pixels or retain too few inliers for reliable visualization. These qualitative observations are consistent with the land-cover-wise AUC results in the table.

\subsection{Robustness to Geometric and Matchability Variation}

\begin{figure}[t]
    \centering
    \includegraphics[width=\linewidth]{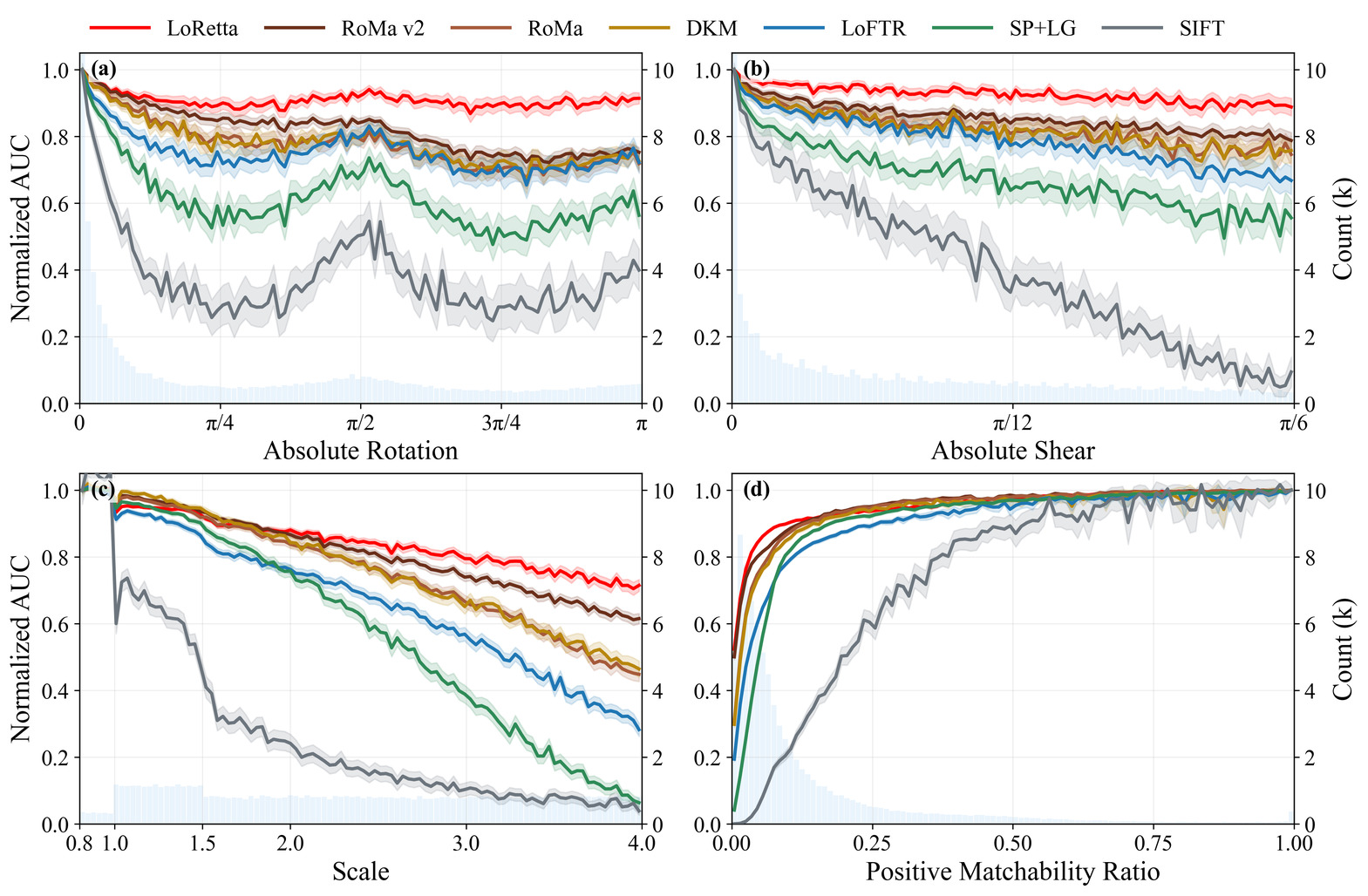}
    \caption{
        \textbf{Robustness under geometric and matchability variation.} The curves show normalized AUC trends with respect to absolute rotation, absolute shear, scale, and positive-matchability ratio. Light-blue bars indicate the number of evaluated samples in each bin.
    }
    \label{fig:exp_robustness}
\end{figure}

Fig.~\ref{fig:exp_robustness} analyzes how performance changes with geometric deformation and matchable-area variation. To emphasize robustness rather than the absolute score of each method, the curves are normalized by an easy reference bin for each factor: near-zero rotation, near-zero shear, the smallest scale range, and the highest positive-matchability ratio. Therefore, a flatter curve indicates that the method preserves more of its reference performance as the test condition moves away from the easy reference bin. These factors correspond directly to the stress cases introduced by LEVIR-GM: full-angle rotations, affine shear, large scale changes, and partially non-matchable regions.

\suppressfloats[t]

Across the geometric variables, LoRetta preserves the highest or near-highest normalized AUC in most bins. The advantage is most visible under larger rotation and shear, where competing methods lose more of their easy-bin performance. Scale variation causes broader degradation across all methods, but LoRetta degrades more slowly than the other dense and semi-dense baselines. For the positive-matchability ratio, performance is lower when only a small fraction of pixels belongs to \(\Omega=\{p\mid \bar M_{1/1}(p)=1\}\), and the gap among strong dense methods becomes smaller as the matchable area increases.

These trends indicate that the four factors stress different aspects of the benchmark. Rotation and shear primarily test whether a matcher can preserve geometric consistency under global affine deformation; scale change reduces shared local details and therefore challenges all methods; and low positive-matchability ratios test whether a method can rely on valid regions when overlap or scene stability is limited. LoRetta's flatter normalized-AUC curves suggest that the localization-and-registration design better preserves dense matching accuracy under both geometric deformation and partial-matchability conditions.

\subsection{Downstream Applications}

Beyond the controlled LEVIR-GM benchmark, we further examine whether LoRetta can serve as a reusable geometric aligner in downstream \rev{applications}. We evaluate two cross-view localization settings, astronaut-to-satellite localization and UAV-to-satellite localization, where the query images are captured by handheld or low-altitude platforms rather than by conventional nadir satellite sensors. These settings test whether a matcher trained for global-scale remote sensing alignment can transfer to practical localization pipelines with stronger viewpoint, scale, appearance, and overlap variations. \rev{Large-image registration further tests whether LoRetta scales to full-size satellite images, and cross-modal registration tests its transferability to other imaging modalities.}

\subsubsection{Astronaut-to-Satellite Localization}

\begin{table}[!t]
    \centering
    \caption{
        Astronaut-to-satellite localization success rate on AIMS using the EarthMatch evaluation procedure. All results except LoRetta are from the original EarthMatch paper. NN and SG denote nearest-neighbor matching and SuperGlue, respectively
    }
    \label{tab:earthmatch_downstream}
    \scriptsize
    \setlength{\tabcolsep}{0.45pt}
    \renewcommand{\arraystretch}{1.06}
    \begin{tabular*}{\columnwidth}{@{\extracolsep{\fill}}C{0.17\columnwidth}*{9}{C{0.075\columnwidth}}@{}}
        \toprule
        \multirow{2}{*}{\raisebox{-3.5ex}{Method}} & \multirow{2}{*}{\raisebox{-3.5ex}{All}} & \multicolumn{4}{c}{Focal (mm)} & \multicolumn{2}{c}{Tilt (\(^{\circ}\))} & \multicolumn{2}{c}{Cloud (\%)} \\
        \cmidrule(lr){3-6}\cmidrule(lr){7-8}\cmidrule(l){9-10}
        & & \(\leq200\) & \(200\)--\(400\) & \(400\)--\(800\) & \(>800\) & \(<40\) & \(\geq40\) & \(<40\) & \(\geq40\) \\
        \midrule
        SIFT-NN        & 70.5 & 64.6 & 74.5 & 64.3 & 82.4  & 74.5 & 58.3 & 76.6 & 57.1 \\
        SP+SG          & 84.0 & 79.3 & 89.1 & 82.1 & 88.2  & 84.8 & 81.7 & 86.8 & 77.9 \\
        SIFT+LG        & 89.3 & 85.4 & 89.1 & 92.9 & 92.2  & 90.8 & 85.0 & 94.0 & 79.2 \\
        SP+LG          & 78.7 & 73.2 & 81.8 & 78.6 & 84.3  & 79.3 & 76.7 & 81.4 & 72.7 \\
        LoFTR          & 72.1 & 72.0 & 78.2 & 60.7 & 78.4  & 71.7 & 73.3 & 71.9 & 72.7 \\
        RoMa           & \underline{93.0} & \underline{90.2} & \underline{90.9} & \underline{92.9} & \textbf{100.0} & \underline{92.4} & \textbf{95.0} & \underline{95.2} & \underline{88.3} \\
        LoRetta (Ours) & \textbf{97.5} & \textbf{96.3} & \textbf{98.2} & \textbf{96.4} & \textbf{100.0} & \textbf{98.4} & \textbf{95.0} & \textbf{99.4} & \textbf{93.5} \\
        \bottomrule
    \end{tabular*}
\end{table}

\begin{figure}[!t]
    \centering
    \includegraphics[width=\linewidth]{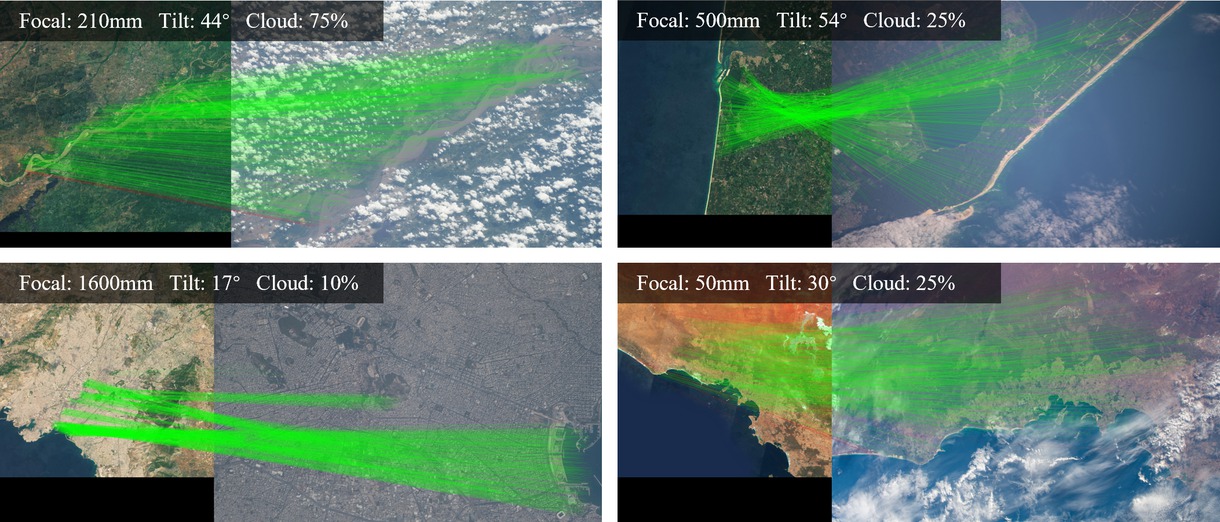}
    \caption{
        \textbf{Astronaut-to-satellite matching examples on AIMS.} Each panel shows a retrieved satellite candidate on the left and the corresponding astronaut query on the right, with LoRetta matches under different focal lengths, camera tilts, and cloud coverages.
    }
    \label{fig:exp_earthmatch}
\end{figure}

EarthMatch~\cite{berton2024earthmatch} addresses fine-grained localization of astronaut photography. Given an astronaut image and satellite candidates retrieved from a large geographic database, it uses image matching to identify the correct candidate and localize the photographed region on the satellite map, and evaluates the full pipeline on the Astronaut Imagery Matching Subset (AIMS) benchmark~\cite{stoken2023findmyastronautphoto}. This setting is a challenging downstream test for global-scale matching because the query image is captured manually from the International Space Station rather than by a fixed nadir-view satellite sensor. The resulting pairs may contain large scale changes, arbitrary in-plane rotations, oblique viewpoints, cloud contamination, radiometric differences, and only partial geographic overlap.

We evaluate LoRetta on AIMS by replacing the matcher in the EarthMatch pipeline while keeping the retrieval candidates and localization protocol unchanged, and compare against the results reported in the original EarthMatch paper. All baseline results in Table~\ref{tab:earthmatch_downstream} are taken from that paper, while only the LoRetta row is produced by our evaluation. Following the original protocol, success is measured as the percentage of localizable astronaut-photo queries that are correctly localized after matching against the retrieved satellite candidates. The table reports the overall success rate and the same acquisition-condition subsets used in EarthMatch, including focal length, camera tilt, and cloud coverage.

\begin{figure}[!t]
    \centering
    \includegraphics[width=\linewidth]{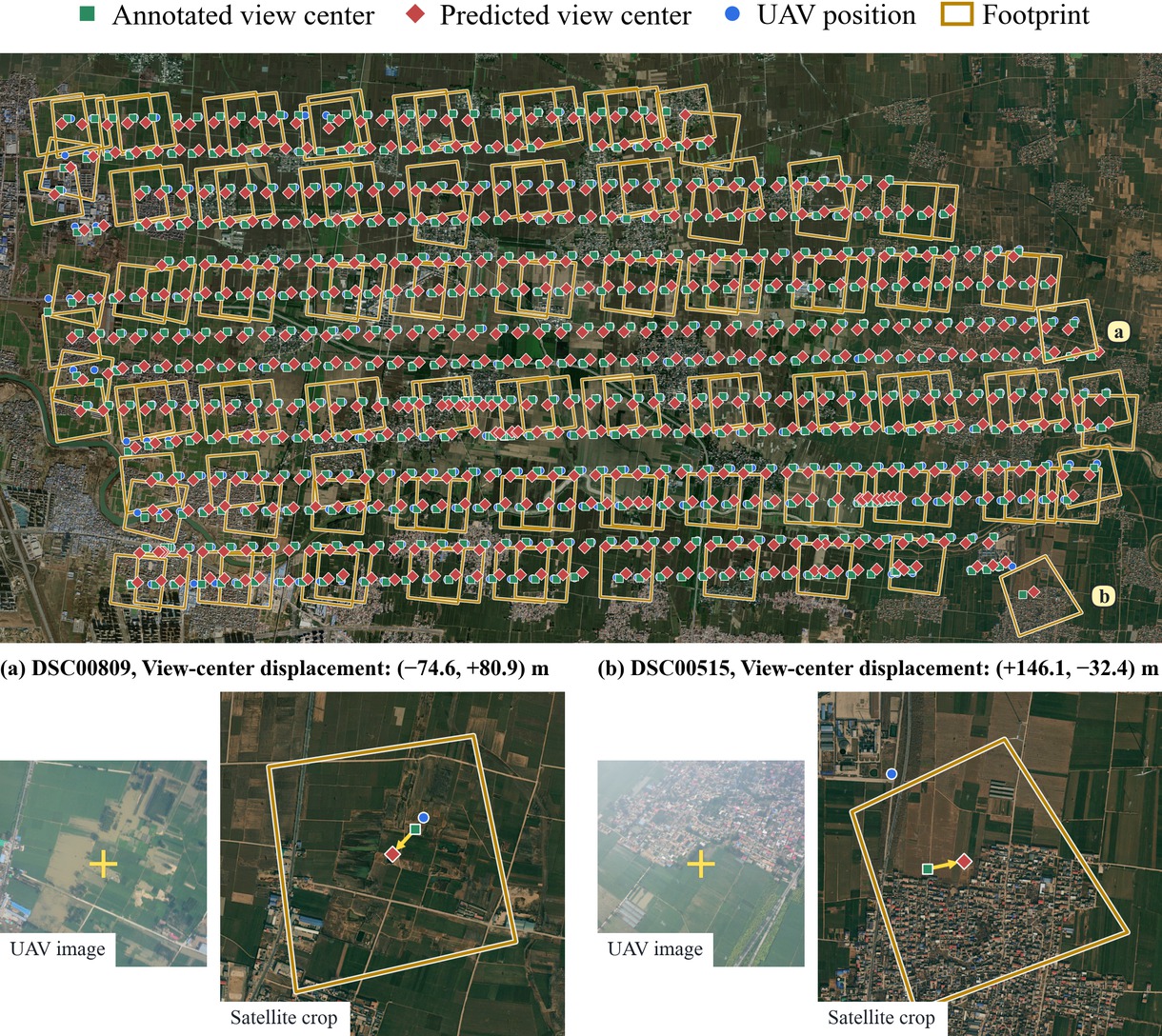}
    \caption{
        \textbf{UAV-to-satellite localization examples on UAVLoc-M3.}
        The CuiJiaQiao overview shows dataset-annotated view centers, LoRetta-predicted view centers, projected UAV positions used for satellite-crop initialization, and estimated homography footprints along the flight trajectories. The \protect\rev{bottom} panels enlarge two examples with their UAV images and corresponding satellite crops; the vectors indicate the annotated-to-predicted view-center displacement.
    }
    \label{fig:exp_uavloc_m3}
\end{figure}

\begin{figure*}[!t]
    \centering
    \includegraphics[width=\textwidth]{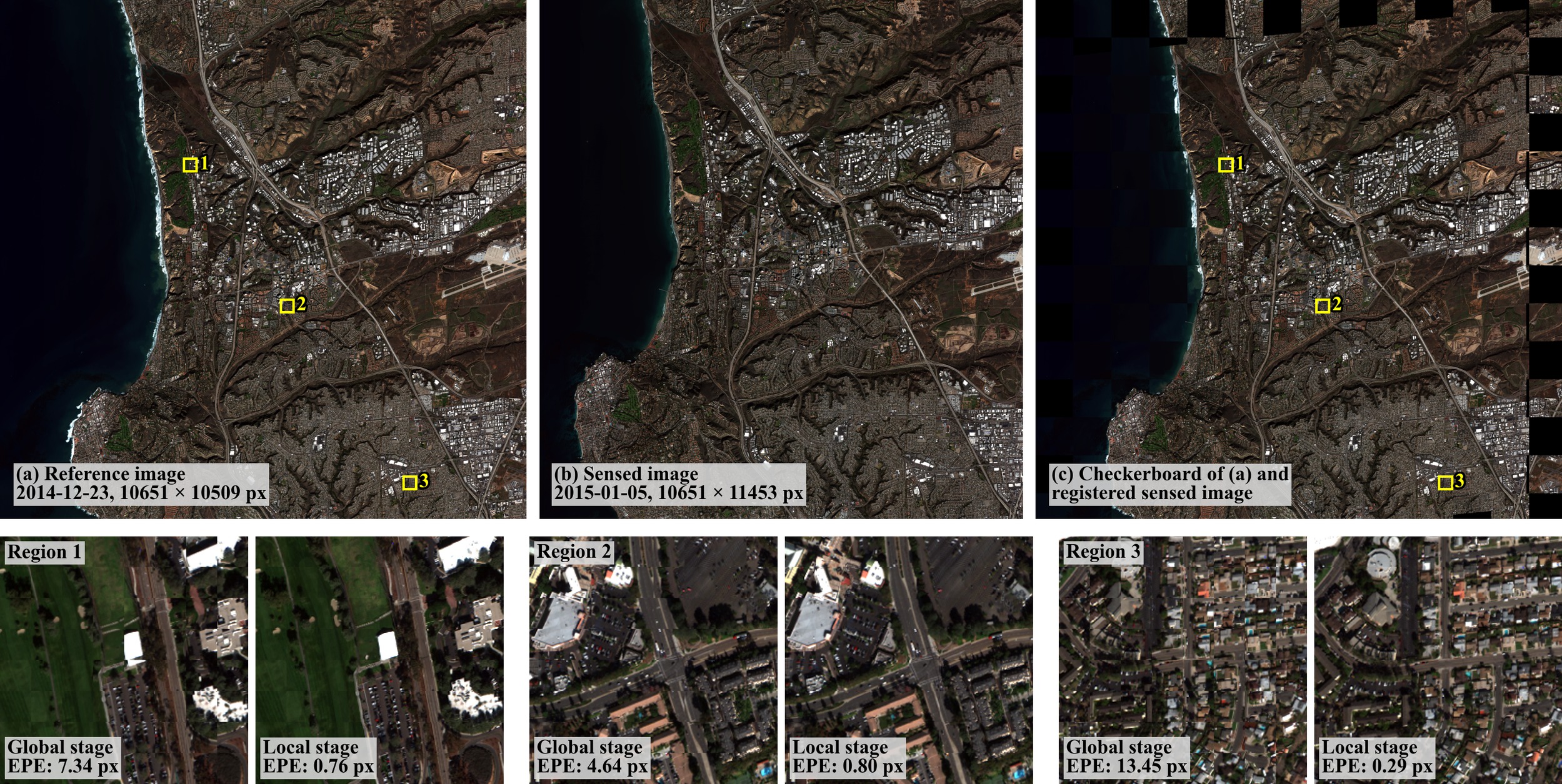}
    \caption{
        \protect\rev{\textbf{Large-image registration example on WorldView-3 images.} Panels (a) and (b) show the reference and sensed images, and (c) shows a checkerboard of (a) and the sensed image registered by LoRetta. The bottom panels enlarge regions 1--3 at full resolution after the global stage (left) and the local stage (right), each labeled with the endpoint error (EPE) at the checkpoint in its center.}
    }
    \label{fig:large_image}
\end{figure*}

LoRetta achieves a \(97.5\%\) overall localization success rate, improving over the strongest reported EarthMatch baseline, RoMa, by \(4.5\) points. The gains are most pronounced in challenging acquisition subsets: LoRetta improves the \(\leq200\) mm focal-length subset from \(90.2\%\) to \(96.3\%\), the \(200\)--\(400\) mm subset from \(90.9\%\) to \(98.2\%\), and the high-cloud subset from \(88.3\%\) to \(93.5\%\). It also matches RoMa's saturated performance on the \(>800\) mm focal-length subset and the high-tilt subset.

Fig.~\ref{fig:exp_earthmatch} further presents representative matching examples across a wide range of focal lengths, camera tilts, and cloud coverages. LoRetta retains spatially consistent correspondences despite substantial scale, viewpoint, radiometric, and occlusion variations, qualitatively supporting the localization results in Table~\ref{tab:earthmatch_downstream}. These results indicate that LoRetta can serve as a reusable matching component in the EarthMatch astronaut-to-satellite localization pipeline.

\subsubsection{UAV-to-Satellite Localization}

We further evaluate LoRetta on UAV-to-satellite localization with UAVLoc-M3~\cite{Nie_UAVLoc_M3_2026}. UAVLoc-M3 contains 4,947 low-altitude UAV images covering three regions: AnYangRiver, Chongmingdao, and CuiJiaQiao, paired with 13 satellite map tiles. The task is to align each UAV image to the corresponding satellite map and estimate the center and spatial footprint of the viewed ground region in satellite-image coordinates. Compared with satellite-to-satellite registration, this setting introduces stronger perspective distortion, scale variation, local scene ambiguity, and partial overlap, because the queries are captured from low altitude and often under non-nadir viewpoints.

To construct each evaluation pair, we recover the UAV platform GPS position from image metadata and project it into satellite-image coordinates using a tile-specific geographic-to-pixel affine transform. This projected position denotes the aircraft location, not the center of the observed ground region, and is used only to center a \(1024\times1024\) satellite crop for coarse initialization. LoRetta then matches the UAV image to the crop, projects the UAV-image center to obtain the predicted view center, and estimates the corresponding homography footprint. This protocol evaluates whether image matching can refine the GPS-based initialization and recover the center and spatial extent of the viewed ground region under low-altitude perspective variation.

Because UAVLoc-M3 provides only coarsely annotated view centers and does not include pixel-accurate registration ground truth, we evaluate geometric matching validity rather than center-level localization error. We report the valid-alignment rate, where a case is considered valid when 1,000 correspondences are retained and the geometric inlier ratio is at least 0.8. Under this criterion, LoRetta achieves \(98.6\%\) on AnYangRiver, \(95.9\%\) on Chongmingdao, and \(98.0\%\) on CuiJiaQiao.

Fig.~\ref{fig:exp_uavloc_m3} visualizes the results on CuiJiaQiao. Across neighboring frames, the predicted view centers and footprints evolve continuously without abrupt spatial jumps. The displacement from the annotated to the predicted view center also exhibits locally consistent changes along the trajectories, which may reflect variations in UAV viewing attitude as well as the limited precision of the center annotations. Together with the high valid-alignment rates, these results indicate that LoRetta provides stable geometric matching for UAV-to-satellite localization.

\subsubsection{Large-Image Registration}\label{sec:large_image}

\rev{We use two unorthorectified WorldView-3 images from the public CORE3D data~\cite{brown2018core3d}, with $10651\times10509$ and $10651\times11453$ pixels, acquired 13 days apart at off-nadir angles of $24.1^\circ$ and $8.2^\circ$. We register them with LoRetta in two stages: a global stage matches $512\times512$ downsampled copies and fits an affine transformation with RANSAC, and a local stage matches 673 native-resolution $512\times512$ tiles after this affine and fuses their dense outputs weighted by matchability. At 30 manually confirmed checkpoints, the RMSE falls from $627.0$~px before registration to $14.0$~px after the global stage and $0.98$~px after the local stage. Fig.~\ref{fig:large_image} visualizes the registration result and three enlarged regions after each stage. With one tile per batch on an RTX 4090, the total inference time of LoRetta is $56.8$~s and its peak GPU memory is $3.8$~GiB. This example indicates that LoRetta can register large satellite images at a modest cost, with memory fixed by the tile size and runtime linear in the number of tiles.}

\subsubsection{Cross-Modal Registration}\label{sec:cross_modal}

\rev{We consider two tasks, Red-to-NIR and optical-to-SAR registration, and score the native dense outputs on the ground-truth overlap. On 294 windows of ISPRS Potsdam orthophotos~\cite{rottensteiner2012isprs} under controlled similarity transforms, Red-to-NIR registration without fine-tuning lowers AUC only from $98.9\%$ to $97.4\%$ compared with Red-to-Red registration. The loss lies mainly at 1--2~px, where PCK@1px drops from $95.8\%$ to $85.9\%$. For optical-to-SAR registration, we fine-tune LoRetta on the OS-dataset~\cite{xiang2020osdataset} with DINOv3 frozen, using the GDROS~\cite{sun2025gdros} split of 2,011 training and 424 test pairs. Fine-tuning raises AUC from $63.1\%$ to $74.0\%$, whereas GDROS, designed for this task, reaches $80.5\%$ with its released weights. These results indicate that LoRetta can transfer to other imaging modalities, although modality-specific designs remain more accurate under large modality gaps.}

\subsection{Ablation Study}

All ablation variants are evaluated with the same TPS-based scoring protocol as the main comparison. When affine localization is removed, the dense registration branch uses an identity affine initialization. When dense registration is removed, the coarse warp field \(W_c\) and matchability map \(M_c\) are interpolated to full resolution and used as the model output. When localization guidance is removed, the affine-localized warping and inverse composition steps are disabled, so the dense branch is no longer explicitly conditioned on the localized affine frame. When matchability sampling is removed, correspondences for geometric estimation are sampled uniformly rather than from high-matchability locations.

\suppressfloats[t]
\begin{table}[t]
    \centering
    \caption{
        Ablation study of LoRetta on LEVIR-GM. PCK thresholds are measured in pixels. Loc., Reg., Guid., and Match. Samp. denote localization, registration, guidance, and matchability sampling, respectively. \protect\rev{The last four rows vary the training data or the locator threshold $\tau_A$ of the full model}
    }
    \label{tab:ablation}
    \scriptsize
    \setlength{\tabcolsep}{0.45pt}
    \renewcommand{\arraystretch}{1.08}
    \begin{tabular*}{\columnwidth}{@{\extracolsep{\fill}}*{4}{C{0.08\columnwidth}}*{6}{C{0.09\columnwidth}}@{}}
        \toprule
        \multirow{2}{*}{\raisebox{-3.0ex}{\shortstack{Affine\\Loc.}}}
        & \multirow{2}{*}{\raisebox{-3.0ex}{\shortstack{Dense\\Reg.}}}
        & \multirow{2}{*}{\raisebox{-3.0ex}{\shortstack{Loc.\\Guid.}}}
        & \multirow{2}{*}{\raisebox{-3.0ex}{\shortstack{Match.\\Samp.}}}
        & \multicolumn{5}{|c}{PCK (\%) \(\uparrow\)}
        & \multirow{2}{*}{\raisebox{-3.0ex}{\shortstack{AUC\\(\%) \(\uparrow\)}}}                                                                                                                                   \\
        \cmidrule(lr){5-9}
        &                                                      &                &                & \multicolumn{1}{|C{0.09\columnwidth}}{1 px}        & 2 px          & 3 px        & 5 px        & 10 px       &               \\
        \midrule
        \(\times\)     & \(\checkmark\)                                       & --             & \(\checkmark\) & \multicolumn{1}{|C{0.09\columnwidth}}{5.7}           & 17.2          & 27.8          & 42.0          & 57.5          & 38.2          \\
        \(\checkmark\) & \(\times\)                                           & --             & \(\checkmark\) & \multicolumn{1}{|C{0.09\columnwidth}}{14.5}          & 35.9          & 53.2          & 75.1          & 92.8          & 67.0          \\
        \(\checkmark\) & \(\checkmark\)                                       & \(\times\)     & \(\checkmark\) & \multicolumn{1}{|C{0.09\columnwidth}}{25.4}          & 48.1          & 63.8          & 81.4          & 94.2          & 73.4          \\
        \(\checkmark\) & \(\checkmark\)                                       & \(\checkmark\) & \(\times\)     & \multicolumn{1}{|C{0.09\columnwidth}}{35.9}          & 62.6          & 76.9          & 89.0          & 96.0          & 80.9          \\
        \(\checkmark\) & \(\checkmark\)                                       & \(\checkmark\) & \(\checkmark\) & \multicolumn{1}{|C{0.09\columnwidth}}{\textbf{38.7}} & \textbf{67.6} & \textbf{81.3} & \textbf{91.4} & \textbf{96.5} & \textbf{83.3} \\
        \midrule
        \revrow \multicolumn{4}{c}{10\% base pairs} & \multicolumn{1}{|C{0.09\columnwidth}}{28.5} & 56.7 & 72.1 & 84.5 & 91.6 & 76.0 \\
        \revrow \multicolumn{4}{c}{25\% base pairs} & \multicolumn{1}{|C{0.09\columnwidth}}{33.9} & 62.9 & 77.3 & 88.1 & 94.0 & 79.8 \\
        \midrule
        \revrow \multicolumn{4}{c}{\(\tau_A=0.2\)} & \multicolumn{1}{|C{0.09\columnwidth}}{38.2} & 67.0 & 80.8 & 91.1 & 96.3 & 82.9 \\
        \revrow \multicolumn{4}{c}{\(\tau_A=0.5\)} & \multicolumn{1}{|C{0.09\columnwidth}}{38.6} & 67.6 & 81.2 & 91.3 & 96.3 & 83.1 \\
        \bottomrule
    \end{tabular*}
\end{table}

Table~\ref{tab:ablation} verifies the contribution of each component. Removing affine localization causes the largest degradation, reducing AUC from \(83.3\%\) to \(38.2\%\). This confirms that direct dense registration is ineffective for globally unaligned remote sensing pairs with large spatial offsets and limited overlap. Using only affine localization without dense registration reaches \(67.0\%\) AUC, showing that the affine branch provides strong macroscopic alignment but cannot fully recover local residual distortions.

When both branches are enabled but localization guidance is removed, AUC drops to \(73.4\%\). This indicates that dense registration must operate in the affine-aligned frame rather than independently re-estimating the entire warp field. Replacing matchability sampling with uniform sampling reduces AUC to \(80.9\%\), indicating that the matchability map helps select geometrically reliable correspondences in complex multi-temporal scenes. \rev{Ranking the fully matchable pixels of each test pair by predicted matchability confirms this: in $93.7\%$ of the pairs, the $10\%$ with the highest matchability have a lower mean endpoint error than all these pixels, and the median of this error over pairs falls from $1.68$ to $1.18$~px.} The full model performs best across all thresholds, validating the coupled design of affine localization, guided residual registration, and matchability sampling.

\rev{For the $10\%$ and $25\%$ rows of Table~\ref{tab:ablation}, we randomly sample these fractions of the aligned base pairs and train the full model only on the training samples generated from them, keeping the test split unchanged. AUC rises from $76.0\%$ to $79.8\%$ and then to $83.3\%$ with all base pairs, and PCK@1px from $28.5\%$ to $33.9\%$ and $38.7\%$. Accuracy thus keeps improving as more base pairs are used, confirming that LoRetta benefits from the large scale of LEVIR-GM.}

\rev{The last two rows vary the matchability threshold $\tau_A$ of the affine locator at test time, without retraining. Changing $\tau_A$ from $0.3$ to $0.2$ or $0.5$ lowers AUC by at most $0.4$ points. A lower $\tau_A$ admits more low-matchability correspondences into the affine fit, whereas a higher $\tau_A$ leaves more pairs with too few correspondences for a reliable fit. The default $\tau_A=0.3$ is thus a safe choice.}

\section{Conclusion}

This paper addressed global-scale remote sensing image matching from both the method and benchmark perspectives. We introduced LEVIR-GM, a large-scale multi-temporal optical matching dataset with dataset-native matchability labels, broad geographic coverage, and resolutions ranging from sub-meter to kilometer scale. We proposed LoRetta, a localization-and-registration dense matching model that combines matchability-weighted affine localization, localization-guided dense registration, and matchability-aware supervision.

Experiments on LEVIR-GM show that LoRetta improves the accuracy-efficiency trade-off over representative sparse, semi-dense, and dense matchers, especially under strict pixel-error thresholds, large affine deformation, and partial matchability. \rev{The land-cover-wise analysis, robustness evaluation, and ablation study} further verify that affine localization, guided residual registration, and matchability sampling are all important to stable dense alignment in complex earth observation scenes. \rev{Experiments on downstream applications} also indicate that LoRetta can serve as a reusable geometric aligner beyond the benchmark. 

\rev{Several directions remain open for future work. Although LoRetta can transfer to other imaging modalities, it still trails modality-specific designs under large modality gaps, which calls for more modality-invariant representations. LEVIR-GM could also be extended with medium-resolution imagery, such as Sentinel-2 and Landsat data, to bridge its high- and low-resolution levels. Finally, strongly non-planar scenes depart from the single affine prior and motivate terrain-aware geometric priors.}

\ifCLASSOPTIONcaptionsoff
    \newpage
\fi

\bibliographystyle{IEEEtran}
\bibliography{references}

\begin{IEEEbiography}[{\includegraphics[width=1in,height=1.25in,clip,keepaspectratio]{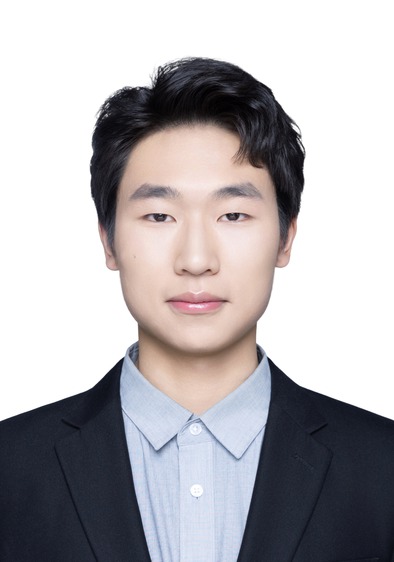}}]{Siwei Yu}
    received his B.S. degree from the School of Astronautics, Beihang University, in 2025. He is currently working toward the Ph.D. degree in the Department of Aerospace Intelligent Science and Technology, School of Astronautics, Beihang University.
    His research interests include deep learning and remote sensing image registration.
\end{IEEEbiography}

\begin{IEEEbiography}[{\includegraphics[width=1in,height=1.25in,clip,keepaspectratio]{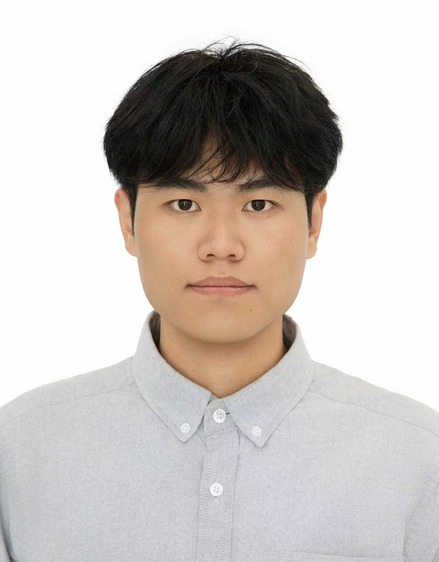}}]{Han Guo}
    received his B.S. degree from the School of Electronic Information Engineering, Inner Mongolia University, in 2024. He is currently working toward the Ph.D. degree in the Department of Aerospace Intelligent Science and Technology, School of Astronautics, Beihang University. His research interests include image processing, machine learning, and pattern recognition.
\end{IEEEbiography}

\begin{IEEEbiography}[{\includegraphics[width=1in,height=1.25in,clip,keepaspectratio]{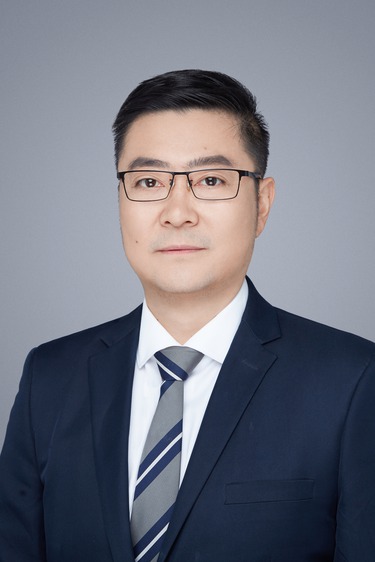}}]{Zhenwei Shi}
    (Senior Member, IEEE) is currently a Professor and Dean of the Department of Aerospace Intelligent Science and Technology, School of Astronautics, Beihang University. He has authored or co-authored over 300 scientific articles in refereed journals and proceedings, including the IEEE Transactions on Pattern Analysis and Machine Intelligence, the IEEE Transactions on Image Processing, the IEEE Transactions on Geoscience and Remote Sensing, the IEEE Conference on Computer Vision and Pattern Recognition (CVPR), and the IEEE International Conference on Computer Vision (ICCV). His current research interests include remote sensing image processing and analysis, computer vision, pattern recognition, and machine learning.

    Dr. Shi serves as an Editor for the IEEE Transactions on Geoscience and Remote Sensing, Pattern Recognition, Remote Sensing, Infrared Physics and Technology, etc.
    His personal website is \url{http://levir.buaa.edu.cn/}.
\end{IEEEbiography}

\begin{IEEEbiography}[{\includegraphics[width=1in,height=1.25in,clip,keepaspectratio]{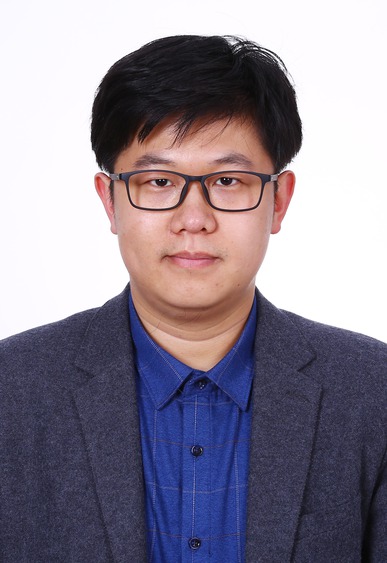}}]{Zhengxia Zou}
    (Senior Member, IEEE) received his B.S. and Ph.D. degrees from Beihang University in 2013 and 2018, respectively. He is currently a Professor at the Department of Aerospace Intelligent Science and Technology, School of Astronautics, Beihang University. During 2018--2021, he was a postdoctoral research fellow at the University of Michigan, Ann Arbor. His research interests include computer vision and related problems in remote sensing. He has published over 100 peer-reviewed papers in top-tier journals and conferences, including Proceedings of the IEEE, Nature Communications, IEEE Transactions on Pattern Analysis and Machine Intelligence, IEEE Transactions on Geoscience and Remote Sensing, and IEEE/CVF Conference on Computer Vision and Pattern Recognition.

    Dr. Zou serves as an Associate Editor for IEEE Transactions on Image Processing and IEEE Transactions on Geoscience and Remote Sensing. His personal website is \url{https://zhengxiazou.github.io/}.
\end{IEEEbiography}

\end{document}